\documentclass{article} 
\usepackage{colm2025_conference}

\usepackage{amsmath,amsfonts,bm}

\def\eqref#1{equation~\ref{#1}}

\def\1{\bm{1}}

\DeclareMathAlphabet{\mathsfit}{\encodingdefault}{\sfdefault}{m}{sl}
\SetMathAlphabet{\mathsfit}{bold}{\encodingdefault}{\sfdefault}{bx}{n}

\usepackage{microtype}
\usepackage{hyperref}
\usepackage{url}
\usepackage{booktabs}
\usepackage{multirow}
\usepackage{lineno}
\usepackage{titlesec}
\usepackage[most,breakable]{tcolorbox}
\usepackage{caption}
\usepackage{xspace}

\definecolor{darkblue}{rgb}{0, 0, 0.5}
\hypersetup{colorlinks=true, citecolor=darkblue, linkcolor=darkblue, urlcolor=darkblue}
\usepackage{enumitem}
\usepackage{graphicx}
\usepackage{subcaption}
\usepackage{colortbl}
\usepackage{amsmath}
\usepackage[capitalise,noabbrev]{cleveref}
\usepackage{mathtools}
\usepackage{siunitx}    
\usepackage{makecell,threeparttable}
\usepackage{algorithm}
\usepackage{algpseudocode}
\usepackage{array}
\usepackage{wrapfig}
\usepackage{svg}
\usepackage{tabularx}
\usepackage{listings}
\usepackage[normalem]{ulem}
\usepackage{soul}

\crefname{section}{Section}{Sections}
\Crefname{section}{Section}{Sections}
\crefrangelabelformat{section}{#3#1--#4#2}

\definecolor{richpurple}{RGB}{75,46,131}

\newcommand{\blackcomic}[1]{%
  {\color{black}\selectfont #1}%
}
\renewcommand{\thefootnote}{\fnsymbol{footnote}}

\newcommand\eat[1]{}

\title{
  \blackcomic{\textbf{From Implicit Weights to Explicit Rubrics}: A Training-Free Framework for Reward Modeling}
}

\author{
\mbox{Lipeng Xie$^{1\ast}$},
\mbox{Sen Huang$^{1\ast}$}, 
\mbox{Zhuo Zhang$^{1\ast}$}, 
\mbox{Anni Zou$^{1}$}, 
\mbox{Yunpeng Zhai$^{1}$}, 
\mbox{Dingchao Ren$^{2}$}, 
\mbox{Kezun Zhang$^{2}$},
\mbox{Haoyuan Hu$^{2}$}, 
\mbox{Boyin Liu$^{1}$}, 
\mbox{Haoran Chen$^{1}$}, 
\mbox{Zhaoyang Liu$^{1\dagger}$}, 
\mbox{Bolin Ding$^{1}$} \\
[1em]
$^1$Tongyi Lab\includegraphics[height=12pt]{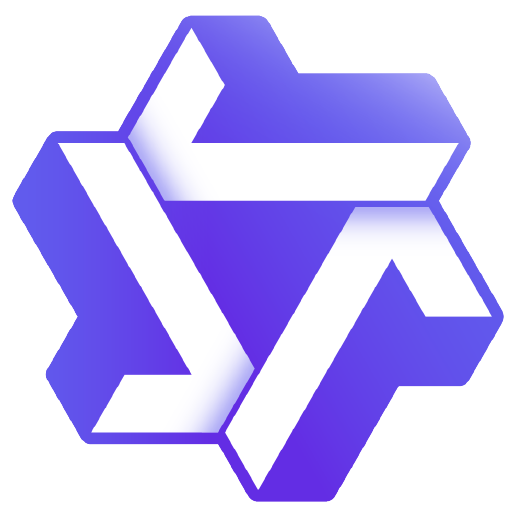}, Alibaba Group \quad $^2$Ant Group \\
[1em]
\texttt{\{xielipeng.xlp, huangsen.huang, zz297429, zouanni.zan, zhaiyunpeng.zyp,} \\
\texttt{liuboyin.lby, congling.chr, jingmu.lzy, bolin.ding\}@alibaba-inc.com} \\
\texttt{\{rendingchao.rdc, kezun.zkz, haoyuan.huhy\}@antgroup.com}
}

\begin{document}

\maketitle

\begingroup
  \renewcommand\thefootnote{}%
  \footnotetext{%
    \begin{tabular}{@{}l@{\hspace{0.4em}}l@{}}
      $^{*}$ & Equal contribution.\\
      $^{\dagger}$ & Corresponding authors.
    \end{tabular}%
  }%
\endgroup

\begin{abstract}
Conventional reward modeling relies on gradient descent over neural weights, creating opaque, data-hungry ``black boxes.'' We propose a paradigm shift from implicit to explicit reward parameterization, recasting optimization from continuous weight spaces to the discrete space of natural language rubrics. We introduce a training-free framework based on iterative rubric learning: it locally induces discriminative criteria via verification-driven refinement, and globally compresses the candidate criteria pool into a compact core set by maximizing an information-theoretic coding rate objective. We organize the compressed core set into a hierarchical rubric structure—high-level evaluation dimensions supported by concrete verification checks—serving as an interpretable, portable reward function. Empirically, our approach challenges prevailing data scaling assumptions: using only \textbf{70 preference pairs}, our rubric-guided judges outperform fully trained reward models on diverse benchmarks. For instance, Qwen3-8B equipped with our learned rubrics achieves 80.91\% on RewardBench2, surpassing the specialized Skywork-Reward-V2-Qwen3-8B (78.20\%). These results demonstrate that alignment signals are highly compressible and can be effectively captured through explicit symbolic search.
\end{abstract}

\section{Introduction}\label{sec:introduction}

The remarkable success of Large Language Models (LLMs) in aligning with human values is largely attributed to Reinforcement Learning from Human Feedback (RLHF) \citep{ouyang2022training,christiano2017deep}. Within this framework, the Reward Model (RM) serves as the critical proxy for human judgment, guiding policy optimization. However, the prevailing approach relies on implicit reward parameterization, where human preferences are compressed into the opaque neural weights of a black-box model. While effective, this approach faces fundamental limitations: it requires massive datasets to perform reliably~\citep{liu2025skywork}, even though recent studies suggest that alignment can be achieved with minimal, high-quality examples \citep{zhou2023lima}. Furthermore, it remains susceptible to reward hacking due to the lack of explicit constraints~\citep{guo2025reward,gao2023scaling}. The opacity of the learned reward function further hinders interpretability, making it difficult to diagnose the rationale behind model preferences.

\begin{figure}[t]
\centering
\makebox[\linewidth][c]{\includegraphics[width=\linewidth, trim=0 10 0 10, clip]{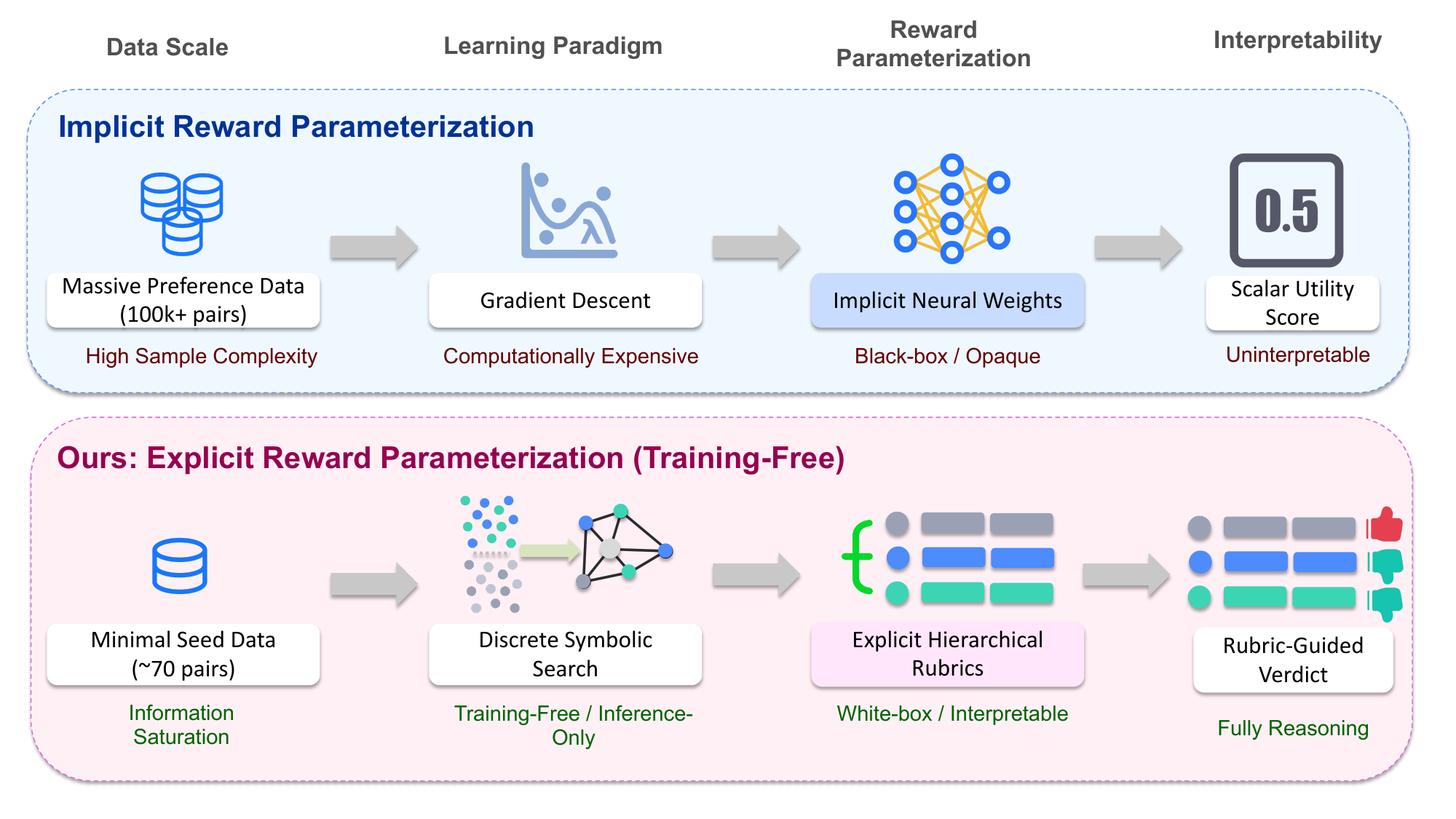}}
    \caption{Paradigm shift from implicit to explicit reward parameterization. \textbf{Top:} Traditional methods rely on data-hungry gradient descent to learn opaque neural weights. \textbf{Bottom:} Our framework induces interpretable evaluation criteria from small data via discrete symbolic search, achieving high efficiency without model training.}
\label{fig:comparison_methods}
\end{figure}

In this work, we pose a fundamental question: \textbf{Can we parameterize the reward function explicitly?} Rather than performing gradient descent over continuous neural weights, we propose optimizing over the \emph{discrete symbolic space} of natural language rubrics—sets of human-readable criteria such as factual accuracy and logical coherence. We reframe alignment not as learning a black-box proxy, but as discovering and refining explicit rubrics directly from human preference data. This paradigm shift from \emph{implicit} to \emph{explicit} parameterization (Figure~\ref{fig:comparison_methods}) provides a transparent alternative to opaque models, enabling interpretable, auditable, and data-efficient alignment.

However, learning effective rubrics from data poses a nontrivial discrete optimization challenge. Existing approaches typically rely either on manual curation~\citep{hashemi2024llm}, which is labor-intensive and non-scalable, or on unverified automated generation~\citep{wang2025autorule,gupta2025carmo}, which often yields noisy and redundant criteria. To address these limitations, we introduce a training-free framework centered on iterative rubric learning:
\begin{itemize}[leftmargin=*, topsep=0pt, partopsep=0pt]
    \item \textbf{Local Induction} treats single-sample learning as active hypothesis testing. We employ a verification-driven refinement loop to \emph{induce} discriminative criteria that can provably explain observed preferences, ensuring high validity and discriminative power at the instance level.
    \item \textbf{Global Compression} treats dataset-level learning as information compression. We map induced rubrics to a semantic embedding space and distill a core set by maximizing the \emph{coding rate}~\citep{yu2020learning}, an information-theoretic objective that explicitly penalizes redundancy while promoting diversity. This effectively \emph{compresses} thousands of local rubrics into a minimal, high-coverage rubric set.
\end{itemize}

The resulting framework produces a hierarchically organized rubric set, structured as high-level evaluation dimensions supported by concrete verification criteria, serving as a portable, interpretable reward model. Crucially, our approach is training-free, requiring no gradient updates and relying solely on inference. Our investigation demonstrates that this symbolic approach offers a competitive alternative to conventional neural methods, showing that high-quality preference alignment is achievable without massive datasets or extensive training.

Our primary contributions are:
\begin{itemize}[leftmargin=*, topsep=0pt, partopsep=0pt]
    \item \textbf{A Paradigm Shift to Explicit Parameterization.} We formulate reward modeling as a search problem in the discrete space of natural language rubrics, offering a transparent, auditable alternative to black-box neural reward models.
    \item \textbf{Iterative Rubric Learning.} We propose a novel algorithm that alternates between local verification-driven induction and global information-theoretic compression, enabling the discovery of high-quality criteria from minimal data.
    \item \textbf{Competitive Performance with High Efficiency.} We demonstrate that our framework matches or exceeds fully trained reward models using orders of magnitude less data. For instance, with only \textbf{70 preference pairs}, our rubric-guided judge (Qwen3-8B) achieves 80.91\% on RewardBench2, surpassing the specialized Skywork-Reward-V2-Qwen3-8B (78.20\%), which is trained on 26 million preference pairs using 64 H800 GPUs. This represents an approximately 370,000$\times$ reduction in data requirements, suggesting that alignment signals are highly compressible and can be captured through explicit symbolic search.
\end{itemize}

\section{Related Work}\label{sec:related_work}

\paragraph{LLM-as-a-Judge with Latent Criteria.} The ``LLM-as-a-Judge'' paradigm approximates human judgments using strong reasoning models~\citep{zheng2023judging,li2023generative}. However, standard approaches predominantly rely on unstructured prompts, where evaluation criteria remain latent within the model weights rather than being explicitly defined. This opacity often leads to inconsistent judgments~\citep{wang2024large,zhu2024starling}. While specialized evaluators like Prometheus~\citep{kim2023prometheus} and reasoning-based methods like RM-R1~\citep{chen2025rm} improve interpretability via fine-grained feedback or Chain-of-Rubrics (CoR), they typically require extensive training (distillation or RL) and lack mechanisms to explicitly decouple the criteria from the model's reasoning process.

\paragraph{Explicit Reward Parameterization.} To enhance reliability and transparency, recent work moves towards parameterizing rewards \emph{explicitly}. Training-based methods like DeepSeek-GRM~\citep{guo2025deepseek,liu2025inference} and R3~\citep{anugraha2025r3} fine-tune models to generate critiques or adapt to provided rubrics. However, they incur high training costs and often assume external rubrics without verifying their quality. Conversely, static methods relying on expert-curated rubrics~\citep{arora2025healthbench,mu2024rule} suffer from poor scalability. Our work bridges this gap by inferring high-quality explicit parameters directly from data, combining the adaptability of learned models with the transparency of rule-based systems, all without gradient updates.

\paragraph{Automated Rubric Discovery.} Closely related is Automated Rubric Discovery. Inverse Constitutional AI (ICAI)~\citep{bai2022constitutional,findeis2024inverse} reverse-engineers principles but often lacks verification, yielding fragmented rules. OpenRubrics~\citep{liu2025openrubrics} improves quality via contrastive generation and consistency filtering, yet requires training on large-scale synthetic data (35.7k pairs) without explicit redundancy control. We advance this by formulating rubric discovery as an iterative learning process. By alternating between verification-driven induction and information-theoretic compression, our framework achieves competitive performance with only 70 preference pairs (a 500$\times$ reduction compared to OpenRubrics) while ensuring the learned rubrics are both discriminatively valid and non-redundant.

\section{Methodology}\label{sec:methods}

\paragraph{Overview.}
This section details our training-free framework for iterative rubric learning. We first formalize the shift from continuous to discrete reward parameterization (Sec.~\ref{sec:formulation}). We then present our core algorithm, which alternates between \emph{local induction} to extract granular criteria and \emph{global compression} to distill a compact core set via information-theoretic optimization. Finally, we introduce quantitative metrics to analyze the quality and non-redundancy of the learned criteria (Sec.~\ref{sec:rubric_analysis}).

\begin{figure*}[!htbp]
  \centering
  \includegraphics[width=0.95\textwidth]{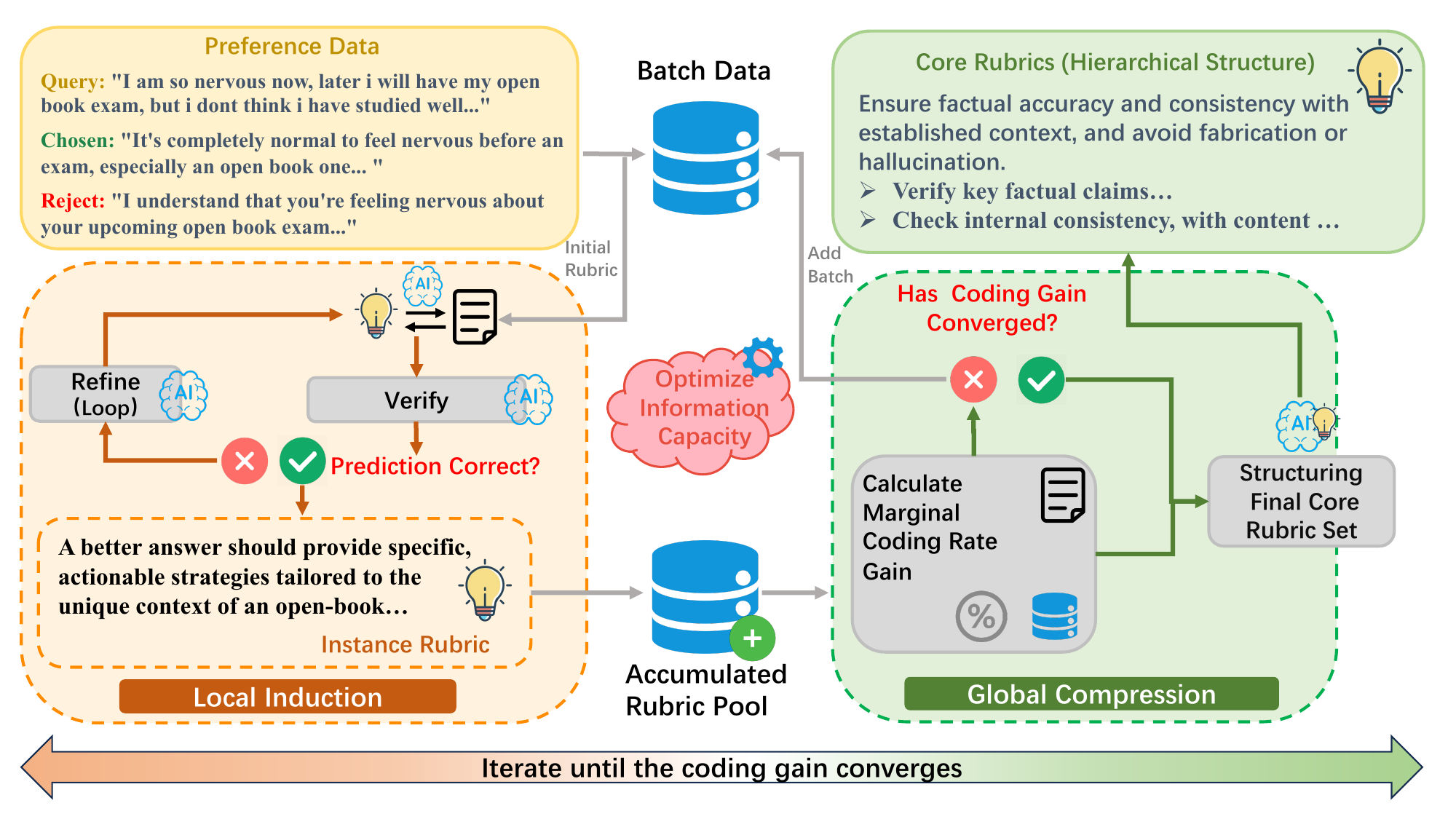}
  \caption{Overview of our iterative rubric learning framework. The learning process alternates between two operations: \textbf{Local Induction} extracts candidate rubrics tailored to individual preference pairs via verification-driven hypothesis refinement, and \textbf{Global Compression} distills the accumulated pool into a compact, high-coverage core set by maximizing an information-theoretic criterion (coding rate). This iterative process continues until convergence.}
  \label{fig:methodology_overview}
\end{figure*}

\subsection{Formulation}\label{sec:formulation}

We formulate preference learning as finding the optimal parameters of a probabilistic preference model. Given a dataset $\mathcal{D} = \{(x_i, y_i^+, y_i^-)\}_{i=1}^N$, the goal is to learn a model $P_\Phi$ that assigns higher likelihood to preferred responses. The critical design choice lies in \emph{the form of the parameters $\Phi$}.

\paragraph{Implicit Parameterization (Bradley-Terry Model).}
The conventional approach parameterizes the preference model with neural network weights $\Phi = \theta \in \mathbb{R}^d$. Under the Bradley-Terry framework~\citep{bradley1952rank}:
\begin{equation} 
\label{eq:bt_model} 
    P_\theta(y^+ \succ y^- | x) = \sigma(r_\theta(x, y^+) - r_\theta(x, y^-)),
\end{equation} 
where $r_\theta$ is a learned scalar reward function. The optimal parameters are obtained via:
\begin{equation} 
\label{eq:bt_optimization} 
    \theta^* = \arg\max_{\theta \in \mathbb{R}^d} \sum_{i=1}^N \log P_\theta(y_i^+ \succ y_i^- | x_i).
\end{equation} 
While effective, this yields an opaque ``black box'' that offers limited interpretability.

\paragraph{Explicit Parameterization (Rubric Learning).}
We propose an alternative: parameterizing the preference model with a natural language rubric set $\Phi = R \subset \mathcal{V}$, where $\mathcal{V}$ is the space of all possible rubrics. Formally, the preference model becomes:
\begin{equation} 
\label{eq:rubric_prob} 
    P_R(y^+ \succ y^- | x) = \text{LLM}_{\text{judge}}(y^+ \succ y^- \mid x, y^+, y^-, R),
\end{equation} 
where $\text{LLM}_{\text{judge}}$ is a prompt-guided evaluator conditioned on $R$ that outputs a binary preference decision. Analogous to Eq.~\ref{eq:bt_optimization}, the idealized objective is:
\begin{equation} 
\label{eq:rubric_optimization} 
    R^* = \arg\max_{R \subset \mathcal{V}} \sum_{i=1}^N \log P_R(y_i^+ \succ y_i^- | x_i).
\end{equation} 
However, directly optimizing Eq.~\ref{eq:rubric_optimization} is intractable: the space $\mathcal{V}$ is discrete and combinatorially large, precluding gradient-based methods. We therefore introduce two relaxations:
\begin{enumerate}[leftmargin=*, topsep=2pt, itemsep=1pt]
    \item \textbf{From log-likelihood to 0-1 accuracy.} 
    Since the LLM judge outputs binary decisions rather than calibrated probabilities, we replace the log-likelihood objective with a 0-1 accuracy surrogate that directly counts correct predictions.
    \item \textbf{From global to local-then-global optimization.} Instead of searching over all possible subsets $R \subset \mathcal{V}$, we decompose the problem: first induce candidate rubrics locally from individual samples, then select a compact core set globally.
\end{enumerate}
The resulting objective becomes:
\begin{equation} 
\label{eq:rubric_accuracy} 
R^* \approx \arg\max_{R \subset \mathcal{R}_{\text{pool}}} \sum_{i=1}^N \mathbb{I}[\text{eval}_R(x_i, y_i^+, y_i^-) = \text{correct}],
\end{equation} 
where $\mathcal{R}_{\text{pool}}$ is the candidate pool generated by local induction. We address this optimization through an iterative algorithm that alternates between \emph{local induction} and \emph{global compression}, refining the rubric set until convergence.

\paragraph{Rubric Representation and Grounding.}
The optimization in Eq.~\ref{eq:rubric_accuracy} aligns rubric selection with preference recovery, but with a critical distinction: the parameterization is \emph{explicit}. Unlike opaque neural weights, these criteria are inherently interpretable and editable. Crucially, even instance-specific details within rubrics serve as \emph{concrete anchors}, guiding the judge model's evaluation via analogical reasoning~\citep{webb2023emergent} rather than superficial keyword matching.
 
\subsection{Iterative Rubric Learning}\label{sec:method_learning}

Our learning algorithm can be understood through an information-theoretic lens as an \emph{induce-then-compress} process. Local induction induces discriminative rubrics from individual preference pairs. Global compression then compresses the accumulated rubrics into a minimal set that preserves maximal information, as measured by the coding rate. This unified perspective aims to maintain high fidelity to the training preferences while ensuring applicability to unseen queries. We now describe each operation in detail.

\paragraph{Local Induction.}
The goal of this step is to induce an explicit rubric that captures the discriminative information embedded in each preference pair. For each instance $(x_i, y_i^+, y_i^-)$, we find a rubric $R_i^*$ that explains \emph{why} the preferred response is better. Since the rubric space $\mathcal{V}$ is discrete, we employ an iterative hypothesis refinement strategy. The process initializes a candidate $R_i^{(0)} = \phi_{\text{init}}(x_i, y_i^+, y_i^-)$ by extracting contrastive features from the response pair. At each step $t$, we verify whether $R_i^{(t)}$ correctly identifies the preferred response:
\begin{equation}\label{eq:evaluate_step}
    \text{verify}(R_i^{(t)}) = \mathbb{I}\left[ \text{eval}_{R_i^{(t)}}(x_i, y_i^+, y_i^-) = \text{correct} \right].
\end{equation}
If verification fails, we apply a refinement update $R_i^{(t+1)} = \textsc{Refine}(R_i^{(t)}, x_i, y_i^+, y_i^-)$. The operator $\textsc{Refine}$ leverages the LLM's reasoning capability to revise the rubric toward one that better explains the observed preference. This adapts the paradigm of iterative self-correction~\citep{madaan2023self, shinn2023reflexion} to the novel context of discrete reward parameter search. This continues until verification succeeds or a maximum of $E_{\max}$ steps. The induced rubrics are accumulated into a candidate pool $\mathcal{R}_{\text{pool}}$.

\paragraph{Global Compression.}
Global Compression distills the noisy, redundant candidate pool $\mathcal{R}_{\text{pool}}$ into a concise core set $R_{\text{core}}$. The raw pool typically contains overlapping or conflicting criteria extracted from diverse queries. Naive concatenation of these rubrics not only inflates inference costs but also introduces reasoning noise that destabilizes the judge model. Our objective is therefore to select a minimal subset that maximizes semantic coverage within a fixed token budget.

While clustering methods (e.g., $k$-means) are common for data reduction, they optimize for intra-cluster compactness rather than global diversity. We instead adopt the coding rate~\citep{yu2020learning}, an information-theoretic objective that measures the volume of the subspace spanned by the rubric embeddings. This metric directly quantifies the \emph{information capacity} of the set, prioritizing rubrics that are semantically distinct.

We model rubrics as vectors on a low-dimensional manifold and define the coding rate as the rate-distortion cost of encoding the embedding matrix $\mathbf{E}_R \in \mathbb{R}^{d \times |R|}$ with tolerance $\varepsilon$:
\begin{equation}\label{eq:coding_rate_aggregation}
    \mathcal{C}(\mathbf{E}_R, \varepsilon) = \frac{1}{2}\log\det\left(\mathbf{I} + \frac{1}{\varepsilon^2 |R|}\mathbf{E}_R^\top\mathbf{E}_R\right).
\end{equation}
Geometrically, maximizing $\mathcal{C}$ equates to maximizing the volume of the parallelepiped spanned by the embeddings. This promotes (1) orthogonality, penalizing redundant (collinear) rubrics, and (2) coverage, favoring rubrics that span new semantic dimensions.

The optimization problem seeks the subset of size at most $m$ that maximizes this rate:
\begin{equation}\label{eq:core_set_optimization}
    R_{\text{core}}^* = \arg\max_{R \subseteq \mathcal{R}_{\text{pool}}, |R| \leq m} \mathcal{C}(\mathbf{E}_R, \varepsilon).   
\end{equation}
We employ a greedy algorithm that iteratively selects the rubric with the highest marginal gain $\Delta(r, R_k)$ (illustrated in Figure~\ref{fig:selection_analysis}):
\begin{equation}\label{eq:greedy_selection}
\begin{aligned}
\Delta(r, R_k) &= \mathcal{C}(\mathbf{E}_{R_k \cup \{r\}}, \varepsilon) - \mathcal{C}(\mathbf{E}_{R_k}, \varepsilon), \\
r_{k+1} &= \arg\max_{r \in \mathcal{R}_{\text{pool}} \setminus R_k} \Delta(r, R_k).
\end{aligned}
\end{equation}
The process terminates when the marginal gain drops below a threshold $\tau_{\min}$. The normalization term $1/|R|$ in Eq.~\ref{eq:coding_rate_aggregation} penalizes redundant additions, allowing the marginal gain to become negative and enabling automatic detection of information saturation.

\paragraph{Convergence and Hierarchical Structuring.}
Algorithm~\ref{alg:pipeline} summarizes the complete workflow. We process preference data iteratively in mini-batches of size $B$, alternating between rubric induction and coding-rate-based compression until the core set $R_{\text{core}}$ stabilizes. Although $R_{\text{core}}$ captures the essential information, a flat list of disjoint criteria can fragment the judge's reasoning. To enforce coherence, we apply a final \emph{hierarchical structuring} step that organizes the discrete criteria into $k$ high-level evaluation dimensions anchored by specific verification checks. This structure acts as a cognitive scaffold, ensuring that abstract assessments are consistently supported by concrete, actionable evidence.

\begin{algorithm}[t]
\caption{Iterative Rubric Learning}
\label{alg:pipeline}
\begin{algorithmic}[1]
\State \textbf{Input:} Preference dataset $\mathcal{D}$.
\State \textbf{Output:} Learned rubric set $R^*$.
\State Initialize $R_{\text{core}} \leftarrow \emptyset$.
\For{iteration $t=1, 2, \dots$ \textbf{until} convergence}
    \State Sample mini-batch $\mathcal{B}_t \subset \mathcal{D}$.
    \State \textit{// Local Induction}
    \State $\mathcal{R}_{\text{new}} \leftarrow \bigcup_{(x,y^+,y^-) \in \mathcal{B}_t} \textsc{Induce}(x, y^+, y^-)$
    \State \textit{// Global Compression}
    \State $\mathcal{R}_{\text{pool}} \leftarrow R_{\text{core}} \cup \mathcal{R}_{\text{new}}$
    \State $R_{\text{core}} \leftarrow \textsc{GreedySelect}(\mathcal{R}_{\text{pool}})$ \hfill $\triangleright$ Eq.~\ref{eq:greedy_selection}
\EndFor
\State Organize $R_{\text{core}}$ into hierarchical structure $\rightarrow R^*$.
\State \textbf{return} $R^*$.
\end{algorithmic}
\end{algorithm}

\subsection{A Framework for Rubric Analysis}\label{sec:rubric_analysis}
To ensure the final rubric set is not only performant but also robust, we introduce a quantitative analysis framework. This framework allows us to dissect the utility of each individual rubric within the learned set $R^*$. By evaluating each rubric along three key dimensions (Eq.~\ref{eq:coverage}--\ref{eq:contribution}), we can validate the effectiveness of our selection process.

Let $\mathcal{D}_{\text{test}} = \{(x_i, y_i^+, y_i^-)\}_{i=1}^{N}$ be the held-out test set. We write $e_R^{(i)} = \text{eval}_{R}(x_i, y_i^+, y_i^-)$ to denote the evaluation outcome of rubric set $R$ on sample $i$. For each rubric $r_j \in R^*$, we define:
\begin{itemize}[leftmargin=*, topsep=0pt, partopsep=0pt]
\item \textbf{Coverage:} The proportion of test samples where the rubric provides a discriminative signal (i.e., does not output ``tie''). This metric measures generality.
\begin{equation}\label{eq:coverage}
\text{Coverage}(r_j) = \frac{1}{N} \sum_{i=1}^{N} \mathbb{I}\!\left[e_{\{r_j\}}^{(i)} \neq \text{tie}\right].
\end{equation}
\item \textbf{Precision:} The probability that the rubric's judgment aligns with the ground truth, conditioned on providing a discriminative signal. This measures reliability.
\begin{equation}\label{eq:precision}
\text{Precision}(r_j) = \frac{\sum_{i=1}^{N} \mathbb{I}\!\left[e_{\{r_j\}}^{(i)} = \text{correct}\right]}{\sum_{i=1}^{N} \mathbb{I}\!\left[e_{\{r_j\}}^{(i)} \neq \text{tie}\right]}.
\end{equation}
\item \textbf{Contribution:} The marginal impact of a rubric on the full set's performance, measured by the accuracy drop upon its removal. This quantifies non-redundancy.
\begin{equation}\label{eq:contribution}
\text{Contribution}(r_j) = \text{Acc}(R^*) - \text{Acc}(R^* \setminus \{r_j\}), \quad \text{where } \text{Acc}(R) = \frac{1}{N} \sum_{i=1}^{N} \mathbb{I}\!\left[e_R^{(i)} = \text{correct}\right].
\end{equation}
\end{itemize}
This framework verifies that our method produces a complementary set, balancing general, high-coverage rubrics with specialized, high-precision ones.

\section{Experiments}\label{sec:experiment}

We empirically validate our framework by investigating three key aspects: (1) the competitive performance of explicit rubrics against implicitly parameterized reward models; (2) the data efficiency of the rubric induction process; and (3) the effectiveness of the learned rubrics in guiding downstream policy alignment.

\subsection{Experimental Setup}\label{sec:experimental_setting}

\paragraph{Preference Datasets for Rubric Induction.} We induce rubrics from two distinct sources:
(1) \textbf{HelpSteer3-Preference}~\citep{wang2025helpsteer3}: An open, human-annotated dataset. We use the \textit{General} domain split to evaluate on unseen distributions.
(2) \textbf{UltraFeedback-Binarized}~\citep{cui2023ultrafeedback}: A large-scale synthetic dataset scored by GPT-4, representing AI-feedback distributions.

\paragraph{Baselines.} We benchmark our framework against a comprehensive suite of evaluators:
(1) \textbf{Zero-shot Base Models}: Standard LLM-as-a-Judge without rubric guidance.
(2) \textbf{In-Context Learning (ICL)}: Few-shot prompting with $k=5$ preference demonstrations~\citep{dong2024survey}.
(3) \textbf{Expert-Curated Judges}: Prompts adapted from Arena-Hard~\citep{li2024crowdsourced} and MT-Bench~\citep{zheng2023judging}, representing strong manual prompt engineering baselines. We also compare with the ICAI protocol~\citep{findeis2024inverse}.
(4) \textbf{State-of-the-Art Reward Models}: Fully trained systems including ArmoRM, J1, R3, RM-R1, and Skywork-Reward-V2~\citep{wang2024interpretable,whitehouse2025j1,liu2025skywork}, serving as the performance upper bound.

\paragraph{Evaluation Benchmarks.} We test on four diverse benchmarks: \textbf{RewardBench}~\citep{lambert2025rewardbench}, \textbf{RewardBench2}~\citep{malik2025rewardbench}, \textbf{RM-Bench}~\citep{liu2024rm}, and \textbf{JudgeBench}~\citep{tan2024judgebench}. These cover chat, reasoning, coding, and safety domains.

\paragraph{Implementation Details.}
Our rubric induction converges after processing 70 preference pairs, yielding approximately 60 candidate rubrics that are then structured into 5 hierarchical evaluation dimensions (see Appendix~\ref{sec:rubric_collections} for complete rubric sets). For downstream alignment, we finetune Qwen2.5-7B-Instruct via DPO~\citep{rafailov2023direct} on WildChat~\citep{zhao2024wildchat} prompts labeled by our rubric-guided judge. Comprehensive details regarding backbone models, evaluation protocols, and other hyperparameters are provided in Appendix~\ref{sec:appendix_experiment_details}.

\subsection{Main Results}\label{sec:main_results}

Table~\ref{tab:main_results_wide} presents the performance of our framework across four benchmarks. We highlight four key findings:

\paragraph{Learned Rubrics Enhance Zero-Shot Judges.}
Explicit rubric guidance consistently improves performance across all judge models and scales. On RewardBench2, rubric guidance yields notable gains: \textbf{+6.54\%} for Qwen3-8B and \textbf{+6.72\%} for Qwen3-32B. Even for the powerful proprietary judge Claude-4-Sonnet, our method achieves 95.81\% on RewardBench (+1.20\%) and 86.90\% on RewardBench2, demonstrating that explicit criteria can unlock reasoning capabilities even in strong models. We observe similar gains for weaker backbones like Llama-3.1-8B (see Appendix~\ref{sec:appendix_backbone_sensitivity}), confirming the method's broad applicability.

\paragraph{Symbolic Search Rivals Neural Training.}
Our training-free framework achieves performance parity with, and frequently surpasses, fully trained reward models. On RewardBench2 and RM-Bench, the Qwen3-8B judge guided by our induced rubrics attains accuracies of 80.91\% and 88.28\%, respectively, outperforming the specialized Skywork-Reward-V2-Qwen3-8B (78.20\% and 82.60\%). These findings indicate that extensive parameter updates are not strictly necessary for high-quality reward modeling; instead, explicit parameterization offers an effective, data-efficient alternative to implicit neural learning.

\paragraph{Robustness Across Data Sources.}
Our framework proves robust to the source of preference data (full results in Table~\ref{tab:appendix_full_results}). Rubrics induced from human annotations (HelpSteer3) and AI feedback (UltraFeedback) achieve comparable average performance: 84.60\% vs. 84.42\% for Qwen3-8B, 86.26\% vs. 85.09\% for Qwen3-32B, and 88.85\% vs. 88.04\% for Claude-4-Sonnet. The small variance ($<$1.2\%) across these fundamentally different feedback distributions demonstrates that our iterative induction process reliably extracts generalizable evaluation criteria regardless of the annotation source.

\paragraph{Cross-Domain Applicability.}
Although our rubrics are learned from general conversational preference data, they transfer effectively to specialized domains. On RM-Bench, we observe positive gains across all major categories: Chat (+6.63\%), Math (+2.60\%), Code (+2.05\%), and Safety-Refuse (+1.96\%), yielding an overall improvement of +2.44\%. While gains vary by domain (e.g., Safety-Response shows marginal changes), the consistent positive transfer suggests that our rubrics capture generalizable evaluation principles. Detailed cross-domain results are provided in Appendix~\ref{sec:appendix_cross_domain}.

\begin{table*}[t]
\centering
\caption{Performance comparison on four key benchmarks (in \%). We compare our framework against zero-shot and prompting baselines across three representative judges. \textbf{Bold} indicates best within each judge block. Full results in Table~\ref{tab:appendix_full_results}.}
\label{tab:main_results_wide}
\resizebox{\textwidth}{!}{%
\begin{tabular}{@{}llccccc@{}}
\toprule
\textbf{Judge} & \textbf{Method} & \textbf{RewardBench} & \textbf{RewardBench2} & \textbf{RM-Bench} & \textbf{JudgeBench} & \textbf{Avg.} \\
\midrule
\multirow{7}{*}{\makecell[l]{Trained RM\\(Reference)}}
& ArmoRM-Llama3-8B & 90.40 & 66.50 & 69.30 & 59.70 & 71.48 \\
& R3-Qwen3-8B & 87.50 & -- & 82.10 & -- & -- \\
& R3-Qwen3-14B & 89.30 & -- & 84.90 & -- & -- \\
& RM-R1-Qwen-32B & 92.90 & -- & 79.10 & -- & -- \\
& RM-R1-DeepSeek-Qwen-32B & 90.90 & -- & 83.90 & -- & -- \\
& Skywork-V2-Qwen3-8B & 93.70 & 78.20 & 82.60 & 73.40 & 81.98 \\
& Skywork-V2-Llama-3.1-8B-40M & 97.80 & 86.50 & 96.00 & 83.40 & 90.93 \\
\midrule
\multirow{7}{*}{Qwen3-8B}
& Zero-shot & 92.93 & 74.37 & 86.90 & 73.14 & 81.84 \\
& ICL (k=5) & 90.18 & 72.57 & 86.83 & 67.71 & 79.32 \\
& Arena-Hard Prompt & 85.63 & 78.93 & 85.88 & \textbf{78.57} & 82.25 \\
& MT-Bench Prompt & 92.56 & 73.40 & 84.84 & 77.14 & 81.99 \\
& ICAI & 93.13 & 79.73 & 88.24 & 76.71 & 84.45 \\
& \cellcolor{gray!12}Ours (HelpSteer3) & \cellcolor{gray!12}\textbf{93.50} & \cellcolor{gray!12}\textbf{80.91} & \cellcolor{gray!12}88.28 & \cellcolor{gray!12}75.71 & \cellcolor{gray!12}\textbf{84.60} \\
& \cellcolor{gray!12}Ours (UltraFeedback) & \cellcolor{gray!12}93.10 & \cellcolor{gray!12}80.54 & \cellcolor{gray!12}\textbf{88.60} & \cellcolor{gray!12}75.43 & \cellcolor{gray!12}84.42 \\
\midrule
\multirow{7}{*}{Qwen3-32B}
& Zero-shot & 92.96 & 75.55 & 85.67 & 75.71 & 82.47 \\
& ICL (k=5) & 90.82 & 75.24 & 85.91 & 74.00 & 81.49 \\
& Arena-Hard Prompt & 89.95 & \textbf{83.22} & 87.09 & 71.43 & 82.92 \\
& MT-Bench Prompt & 91.99 & 74.29 & 80.80 & 75.44 & 80.63 \\
& ICAI & 93.37 & 82.57 & 87.37 & 70.85 & 83.54 \\
& \cellcolor{gray!12}Ours (HelpSteer3) & \cellcolor{gray!12}\textbf{93.80} & \cellcolor{gray!12}82.27 & \cellcolor{gray!12}\textbf{88.11} & \cellcolor{gray!12}\textbf{80.86} & \cellcolor{gray!12}\textbf{86.26} \\
& \cellcolor{gray!12}Ours (UltraFeedback) & \cellcolor{gray!12}93.03 & \cellcolor{gray!12}80.69 & \cellcolor{gray!12}87.50 & \cellcolor{gray!12}79.14 & \cellcolor{gray!12}85.09 \\
\midrule
\multirow{7}{*}{Claude-4-Sonnet}
& Zero-shot & 94.61 & 86.70 & 85.70 & 78.29 & 86.33 \\
& ICL (k=5) & 94.82 & 84.89 & 83.29 & 77.61 & 85.15 \\
& Arena-Hard Prompt & 95.41 & 84.99 & 88.11 & 82.57 & 87.77 \\
& MT-Bench Prompt & 95.04 & 83.27 & 83.84 & 76.57 & 84.68 \\
& ICAI & 94.57 & 86.22 & 89.34 & 80.86 & 87.75 \\
& \cellcolor{gray!12}Ours (HelpSteer3) & \cellcolor{gray!12}\textbf{95.81} & \cellcolor{gray!12}86.90 & \cellcolor{gray!12}\textbf{89.49} & \cellcolor{gray!12}\textbf{83.18} & \cellcolor{gray!12}\textbf{88.85} \\
& \cellcolor{gray!12}Ours (UltraFeedback) & \cellcolor{gray!12}95.04 & \cellcolor{gray!12}\textbf{87.90} & \cellcolor{gray!12}87.50 & \cellcolor{gray!12}81.71 & \cellcolor{gray!12}88.04 \\
\bottomrule
\end{tabular}%
}
\end{table*}

\subsection{Data Efficiency and Convergence Analysis}\label{sec:data_efficiency}

We investigate why our framework requires orders of magnitude less data than training-based methods. Our analysis on HelpSteer3 reveals that high efficiency stems from the low intrinsic dimensionality of human preference criteria.

\paragraph{High-Quality Induction from Few Shots.}
Unlike gradient-based methods that require accumulating statistical evidence from large datasets, our verification-driven induction ensures that \emph{every} single induced rubric is discriminative. As shown in Figure~\ref{fig:query_specific_accuracy}, locally induced rubrics achieve $>90\%$ accuracy within just 3--5 refinement iterations. This means we induce high-quality signal from each preference pair, maximizing the information gain per sample.

\begin{figure}[t]
\centering
\begin{minipage}[b]{0.48\textwidth}
\centering
\includegraphics[width=\textwidth]{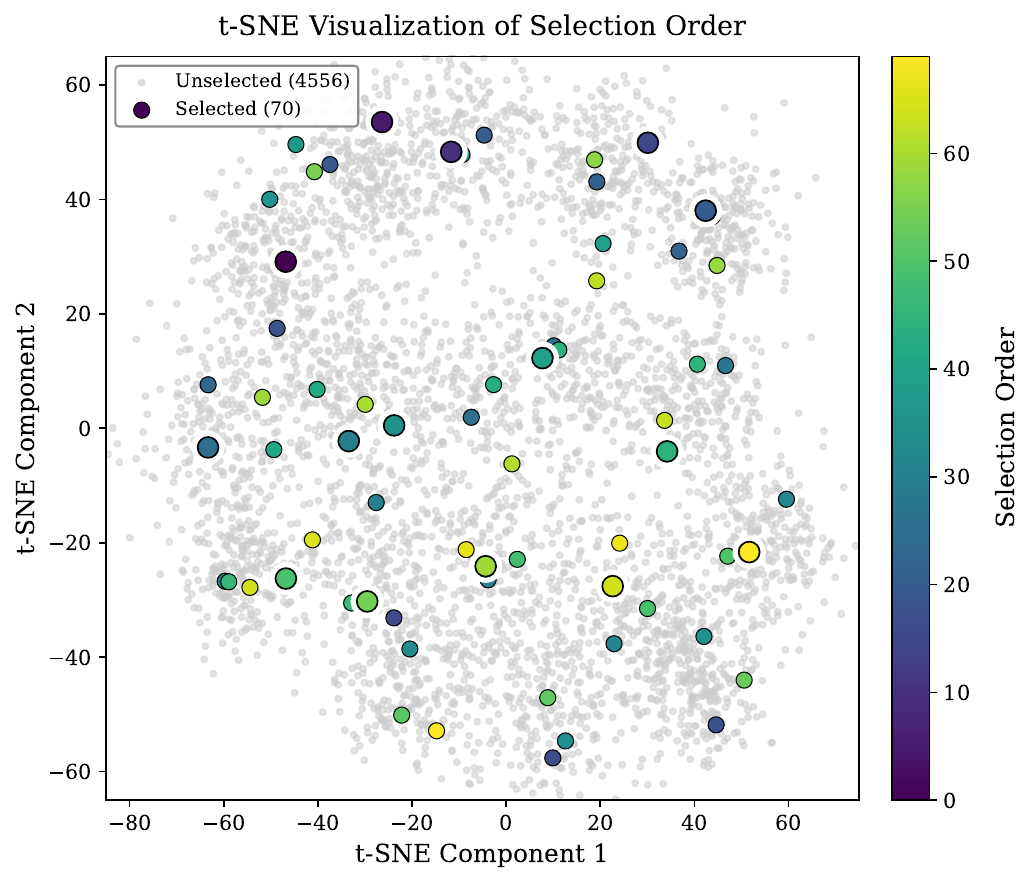}
\subcaption{Rubric Selection in Embedding Space}
\label{fig:selection_order}
\end{minipage}
\hfill
\begin{minipage}[b]{0.48\textwidth}
\centering
\includegraphics[width=\textwidth]{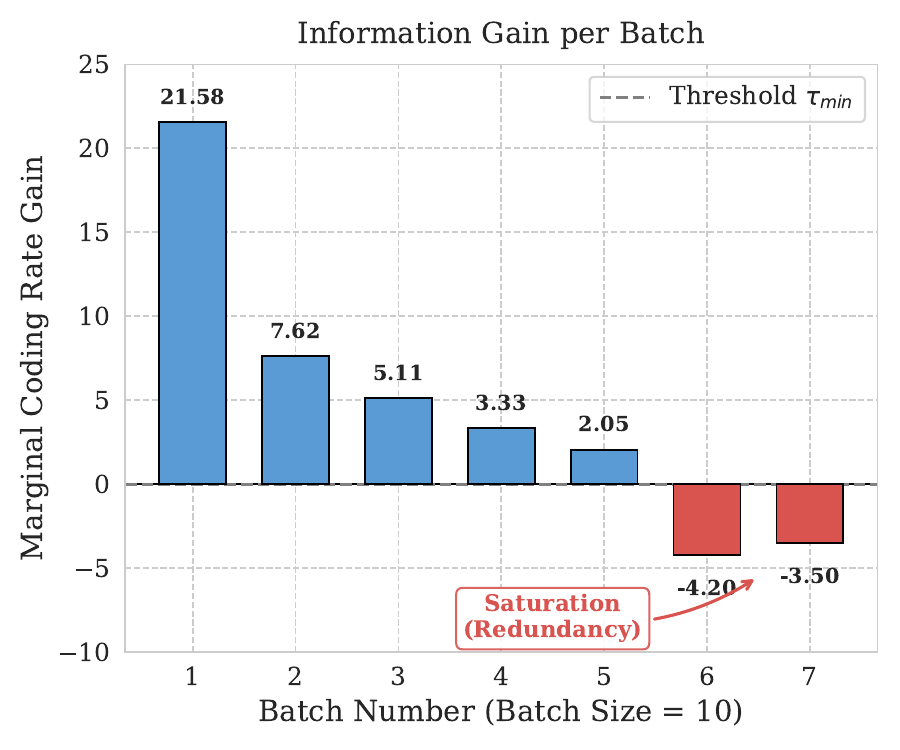}
\subcaption{Marginal Coding Rate Gain per Batch}
\label{fig:batch_increments}
\end{minipage}
\caption{Convergence of rubric compression. (a) t-SNE visualization of the rubric embedding space. Selected rubrics (colored points) sparsely cover the semantic manifold, with early selections (purple) establishing global coverage. (b) Marginal gain in Coding Rate per batch (batch size = 10). The gain becomes negative when the redundancy penalty outweighs the volumetric expansion, indicating information saturation at approx. 60 rubrics.}
\label{fig:selection_analysis}
\end{figure}

\paragraph{Rapid Saturation of Information.}
The core insight is that human preferences are governed by a compact set of latent rubrics. Figure~\ref{fig:selection_analysis} visualizes the rubric compression process. As we greedily add rubrics maximizing the coding rate, the marginal gain diminishes rapidly.
\begin{itemize}[leftmargin=*, topsep=0pt, partopsep=0pt]
    \item \textbf{Global Coverage (Fig.~\ref{fig:selection_order}):} Early selections (purple points) are sparsely distributed across the semantic manifold, quickly establishing global coverage of the preference space.
    \item \textbf{Information Saturation (Fig.~\ref{fig:batch_increments}):} The marginal coding rate gain drops below zero at Batch 6. Following our patience criterion ($p_{\text{patience}}=2$, see Appendix~\ref{sec:appendix_hyperparameters}), the process terminates at Batch 7 (70 pairs). This negative gain indicates that the redundancy penalty outweighs expansion, confirming that the core set has saturated the information content.
\end{itemize}
This rapid saturation explains why 1.5\% of the data is sufficient: the remaining data essentially repeats the same evaluation principles. Full hyperparameter settings are in Appendix~\ref{sec:appendix_hyperparameters}.

\subsection{Ablation Studies}\label{sec:ablation_studies}

We dissect our framework to validate the necessity of its three-stage learning process. Table~\ref{tab:ablation_comprehensive} reports the performance impact of each component on RewardBench2 and RM-Bench.

\paragraph{Efficacy of Local Induction.}
Our \textit{Full Iterative Refinement} outperforms \textit{Single-pass Generation} by +2.43\% on RewardBench2 and +2.04\% on RM-Bench. While \textit{Blind Revision} (without verification) offers some gains on RewardBench2 (+2.14\%), it fails on RM-Bench (-0.28\%). This confirms that only by actively verifying against the preference label can the model induce reliable signals.

\paragraph{Necessity of Global Compression.}
For rubric aggregation, our information-theoretic \textit{Coding Rate Maximization} outperforms \textit{Random Selection} by substantial margins (+3.16\% on RewardBench2). This validates that maximizing the coding rate effectively filters redundancy and enforces semantic diversity, whereas random selection fails to capture the comprehensive preference manifold.

\paragraph{Benefit of Hierarchical Structuring.}
Formatting rubrics into a \textit{Hierarchical Structure} yields superior performance over a \textit{Flat List} (+1.13\% on RewardBench2). This indicates that organizing criteria into broad dimensions anchored by concrete checks provides a more robust cognitive scaffold, helping the judge balance broad evaluation dimensions with concrete verification checks.

\begin{table}[t]
\centering
\caption{Ablation analysis of core components using Qwen3-32B as judge with HelpSteer3 rubrics. We report accuracy (\%) and the performance gap ($\Delta$) relative to the baseline in each group. \textbf{Bold} denotes the configuration used in our framework.}
\label{tab:ablation_comprehensive}
\small
\setlength{\tabcolsep}{3pt}
\begin{tabular}{@{}lcccc@{}}
\toprule
\multirow{2}{*}{\textbf{Method Variant}} & \multicolumn{2}{c}{\textbf{RewardBench2}} & \multicolumn{2}{c}{\textbf{RM-Bench}} \\
\cmidrule(lr){2-3} \cmidrule(l){4-5}
& \textbf{Acc.} & \textbf{$\Delta$} & \textbf{Acc.} & \textbf{$\Delta$} \\
\midrule
\multicolumn{5}{@{}l@{}}{\textit{\textbf{Local Induction Strategy}}} \\
\hspace{3mm}Single-pass Generation & 79.84 & -- & 86.07 & -- \\
\hspace{3mm}Blind Revision (No Verif.) & 81.98 & +2.14 & 85.79 & -0.28\\
\hspace{3mm}\textbf{Full Iterative Refinement} & \textbf{82.27} & \textbf{+2.43} & \textbf{88.11} & \textbf{+2.04} \\
\midrule
\multicolumn{5}{@{}l@{}}{\textit{\textbf{Global Compression Strategy}}} \\
\hspace{3mm}Random Selection & 79.11 & -- & 86.80 & -- \\
\hspace{3mm}\textbf{Coding Rate Maximization} & \textbf{82.27} & \textbf{+3.16} & \textbf{88.11} & \textbf{+1.31} \\
\midrule
\multicolumn{5}{@{}l@{}}{\textit{\textbf{Hierarchical Structuring}}} \\
\hspace{3mm}Flat List (No Structure) & 81.14 & -- & 87.41 & -- \\
\hspace{3mm}\textbf{Hierarchical Structure} & \textbf{82.27} & \textbf{+1.13} & \textbf{88.11} & \textbf{+0.70} \\
\bottomrule
\end{tabular}
\end{table}

\subsection{Application to Policy Optimization}\label{sec:dpo_application}

Beyond evaluation, we investigate the utility of learned rubrics as high-quality training signals for policy optimization. We employ our framework to label preference pairs on WildChat and finetune \textbf{Qwen2.5-7B-Instruct} via DPO. Table~\ref{tab:dpo_results} compares our method against baselines trained on standard rewards, including ArmoRM, and Skywork.

\begin{table}[t]
\centering
\caption{DPO performance comparison on Arena-Hard and AlpacaEval. We report win-rates (\%) against the baseline. \textbf{Style-Ctrl} refers to Style-Controlled win rates for Arena-Hard and Length-Controlled win rates for AlpacaEval. Baseline results ($\dagger$) are sourced from \citet{viswanathan2025checklists}.}
\label{tab:dpo_results}
\small
\setlength{\tabcolsep}{2.5pt} 
\begin{tabularx}{\columnwidth}{@{}>{\raggedright\arraybackslash}X
                                >{\centering\arraybackslash}p{0.14\columnwidth}
                                >{\centering\arraybackslash}p{0.14\columnwidth}
                                >{\centering\arraybackslash}p{0.14\columnwidth}
                                >{\centering\arraybackslash}p{0.14\columnwidth}@{}}
\toprule
& \multicolumn{2}{c}{\textbf{Arena-Hard}} & \multicolumn{2}{c}{\textbf{AlpacaEval}} \\
\cmidrule(lr){2-3} \cmidrule(lr){4-5}
\textbf{Model / Strategy} & \textbf{Vanilla} & \textbf{Style-Ctrl} & \textbf{Vanilla} & \textbf{Style-Ctrl} \\
\midrule
GPT-4-0314 (Reference)$^\dagger$ & 50.0 & 50.0 & 22.1 & 35.3 \\
Qwen2.5-7B-Instruct (Base)$^\dagger$ & 51.3 & 42.8 & 33.5 & 36.2 \\
+ SFT (Distilled)$^\dagger$ & 32.6 & 29.2 & 36.1 & 33.3 \\
\midrule
+ DPO (UltraFeedback)$^\dagger$ & 52.8 & 47.9 & 33.7 & 38.7 \\
+ DPO (AI Judge)$^\dagger$ & 51.0 & 44.4 & 28.8 & 33.4 \\
+ DPO (ArmoRM)$^\dagger$ & 50.8 & 46.4 & 37.6 & 38.1 \\
+ DPO (Skywork)$^\dagger$ & \textbf{55.1} & 50.3 & \textbf{44.8} & 41.5 \\
+ DPO (RLCF)$^\dagger$ & 54.6 & 48.4 & 36.2 & 37.1 \\
\rowcolor{gray!10} + DPO (\textbf{Ours}) & 54.2 & \textbf{57.0} & 42.8 & \textbf{42.2} \\
\bottomrule
\end{tabularx}
\end{table}

The results highlight a key advantage of our approach: mitigation of length bias. While Skywork achieves top vanilla scores (55.1\% Arena-Hard), it drops significantly (-4.8\%) under style-controlled evaluation, suggesting reliance on response verbosity (``length-hacking''). In contrast, our approach exhibits a \textbf{positive gain} (+2.8\%) after style control, achieving the highest controlled win-rate of 57.0\%. This confirms that our explicit rubrics, which directly prioritize qualities like ``Clarity'' and ``Conciseness,'' successfully guide the policy to optimize for genuine response quality rather than superficial length~\citep{dubois2024length}. Similar patterns emerge on AlpacaEval, where our method achieves competitive vanilla performance (42.8\%) while maintaining the highest length-controlled win-rate (42.2\%). These results demonstrate that rubric-guided preference labels provide a more robust training signal for policy optimization.

\section{Conclusion}\label{sec:conclusion}

In this work, we present a paradigm shift in reward modeling, moving from implicit neural weight optimization to \textbf{explicit rubric learning}. By formulating reward modeling as a discrete search process driven by iterative local induction and information-theoretic compression, we demonstrate that high-fidelity judges can be constructed without gradient updates. Our framework distills latent preference signals from as few as 70 samples into interpretable hierarchical rubrics, achieving performance that rivals or exceeds fully trained reward models. This data efficiency challenges the prevailing assumption that alignment requires massive datasets. Furthermore, explicit parameterization offers a transparent alternative to opaque ``black-box'' scoring, grounding LLM alignment in verifiable, human-readable criteria. We hope this work inspires further research into \emph{explicit reward modeling} paradigms, paving the way for more data-efficient, controllable, and trustworthy AI systems.

\bibliography{colm2025_conference}
\bibliographystyle{colm2025_conference}

\clearpage

\appendix
\onecolumn

\section{Limitations}
\label{sec:limitations}
We identify three primary limitations. \textbf{(i) Dependence on Backbone Capabilities:} Our framework acts as a performance multiplier rather than a substitute for base model capability. While it consistently improves weak models (e.g., Llama-3.1-8B; see Appendix~\ref{sec:appendix_backbone_sensitivity}), their absolute performance remains bounded by their inherent reasoning and instruction-following limits compared to stronger backbones. \textbf{(ii) Domain Specialization:} Rubrics distilled from general chat data may not fully capture the nuances of highly specialized domains (e.g., adversarial safety), suggesting a need for domain-targeted induction. \textbf{(iii) Inference Overhead:} The reliance on test-time scaling (e.g., Voting@5) to maximize rubric effectiveness incurs higher computational costs compared to single-pass reward models, presenting a trade-off between judgment precision and inference cost.

\section{Algorithm Hyperparameters}
\label{sec:appendix_hyperparameters}

To ensure reproducibility, we provide detailed definitions of all hyperparameters used in our framework. These parameters control the iterative rubric learning process described in Section~\ref{sec:data_efficiency}.

\begin{table}[h]
\centering
\caption{Hyperparameters used in our iterative rubric learning algorithm.}
\label{tab:hyperparameters}
\small
\renewcommand{\arraystretch}{1.1}
\begin{tabular}{@{}lll@{}}
\toprule
\textbf{Symbol} & \textbf{Value} & \textbf{Description} \\
\midrule
\multicolumn{3}{l}{\textit{Local Induction}} \\
\quad $E_{\text{max}}$ & 10 & Max refinement iterations per preference pair \\
\midrule
\multicolumn{3}{l}{\textit{Global Compression}} \\
\quad Embedding & text-embedding-v1 & Model for encoding rubrics into semantic vectors \\
\quad $B$ & 10 & Batch size (preference pairs per iteration) \\
\quad $\varepsilon$ & 0.5 & Distortion tolerance in coding rate (Eq.~\ref{eq:coding_rate_aggregation}) \\
\quad $\tau_{\text{min}}$ & 0.002 & Min marginal gain to update core set \\
\quad $p_{\text{patience}}$ & 2 & Consecutive low-gain iterations before termination \\
\midrule
\multicolumn{3}{l}{\textit{Hierarchical Structuring}} \\
\quad $k$ & 5 & Number of high-level evaluation dimensions \\
\bottomrule
\end{tabular}
\end{table}

\paragraph{Note on $E_{\text{max}}$:} Unlike training epochs in deep learning, $E_{\text{max}}$ refers to self-correction iterations for a single sample, not passes over the entire dataset. As shown in Figure~\ref{fig:query_specific_accuracy}, convergence typically occurs within 3--5 iterations.

\paragraph{Note on $\varepsilon$:} This parameter controls the ``negative information gain'' phenomenon in Figure~\ref{fig:batch_increments}. As $|R_{\text{core}}|$ increases, the penalty term $\frac{1}{\varepsilon^2 |R_{\text{core}}|}$ in Eq.~\ref{eq:coding_rate_aggregation} eventually outweighs semantic expansion from redundant rubrics, triggering early termination.

\section{Extended Experimental Details}
\label{sec:appendix_experiment_details}

This section provides the technical details necessary for reproducibility, complementing Section~\ref{sec:experimental_setting}.

\paragraph{Backbone Models.}
We employ \textbf{Qwen3-32B}~\citep{yang2025qwen3} as the backbone for iterative rubric learning. For downstream evaluation, we apply the learned rubrics to various judge models, including open-weights systems (Qwen3-8B, 14B, 32B, 235B) and proprietary APIs (Claude-4-Sonnet, GPT-4o), to assess cross-model compatibility. A sensitivity analysis comparing Qwen and Llama backbones is provided in Appendix~\ref{sec:appendix_backbone_sensitivity}.

\paragraph{Evaluation Protocol.}
To ensure fair comparison, all prompt-based methods use identical judge models. We apply a \emph{single fixed rubric set} per source dataset across all benchmarks without task-specific tuning. Voting strategies are standardized as follows:
\begin{itemize}[leftmargin=*, nosep]
    \item \textbf{RewardBench2}: Voting@10 (majority vote).
    \item \textbf{RewardBench \& JudgeBench}: Voting@5.
    \item \textbf{RM-Bench}: Voting@1 (single pass), following the benchmark's default evaluation setting.
\end{itemize}
A test-time scaling analysis is provided in Appendix~\ref{sec:appendix_scaling}.

\paragraph{Inference Configuration.}
All open-weights model inference is performed with the following sampling parameters: \texttt{temperature=0.7}, \texttt{top\_p=0.8}, \texttt{top\_k=20}. We enable ``think mode'' (chain-of-thought) for both rubric induction and evaluation. Proprietary models use standard API endpoints with default parameters.

\paragraph{Compute Accounting.}
We distinguish between \emph{learning cost} and \emph{evaluation cost}:
\begin{itemize}[leftmargin=*, nosep]
    \item \textbf{Learning (One-time):} Processing 70 pairs with up to 10 refinement iterations incurs minimal overhead, negligible compared to the hundreds of GPU-hours required for reward model training.
    \item \textbf{Evaluation (Recurring):} Dominated by the voting budget. For each benchmark, we apply the same voting strategy to all methods, ensuring that observed performance differences reflect rubric quality rather than computational advantage.
\end{itemize}

\paragraph{DPO Training Details.}
For policy optimization (Sec.~\ref{sec:dpo_application}), we finetune \textbf{Qwen2.5-7B-Instruct} on WildChat prompts labeled by our rubric-guided judge (HelpSteer3 source):
\begin{itemize}[leftmargin=*, nosep]
    \item \textbf{Hyperparameters:} 2 epochs, global batch size 1024, max sequence length 2048, learning rate 5e-7 (cosine decay).
    \item \textbf{Infrastructure:} 2 $\times$ 8 H20 nodes.
\end{itemize}

\section{Analysis on Cross-Model Compatibility of Rubrics}
\label{sec:appendix_rubric_analysis}

\begin{figure}[t!]
\centering
\begin{minipage}[t]{0.48\textwidth}
\centering
\includegraphics[width=\textwidth]{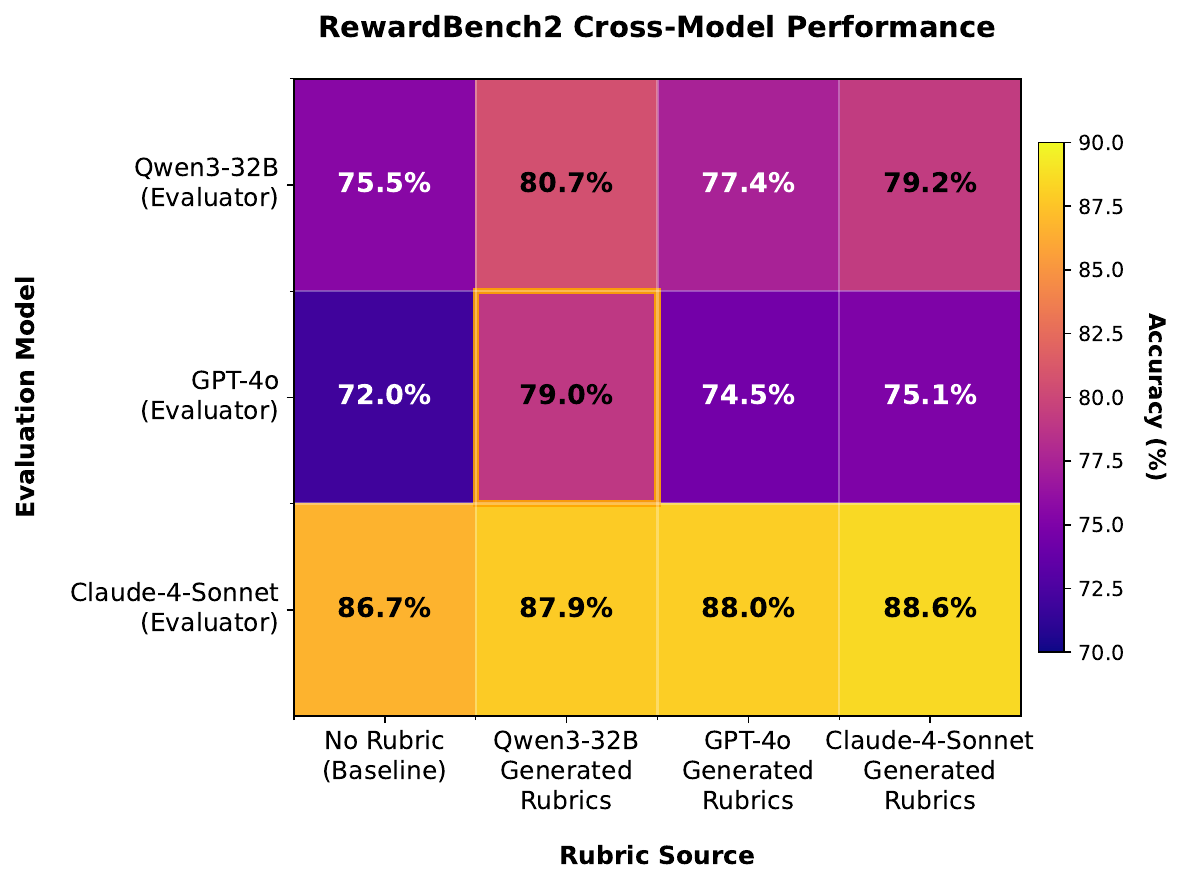}
\subcaption{RewardBench2 Performance}
\label{fig:appendix_rb2_heatmap}
\end{minipage}
\hfill
\begin{minipage}[t]{0.48\textwidth}
\centering
\includegraphics[width=\textwidth]{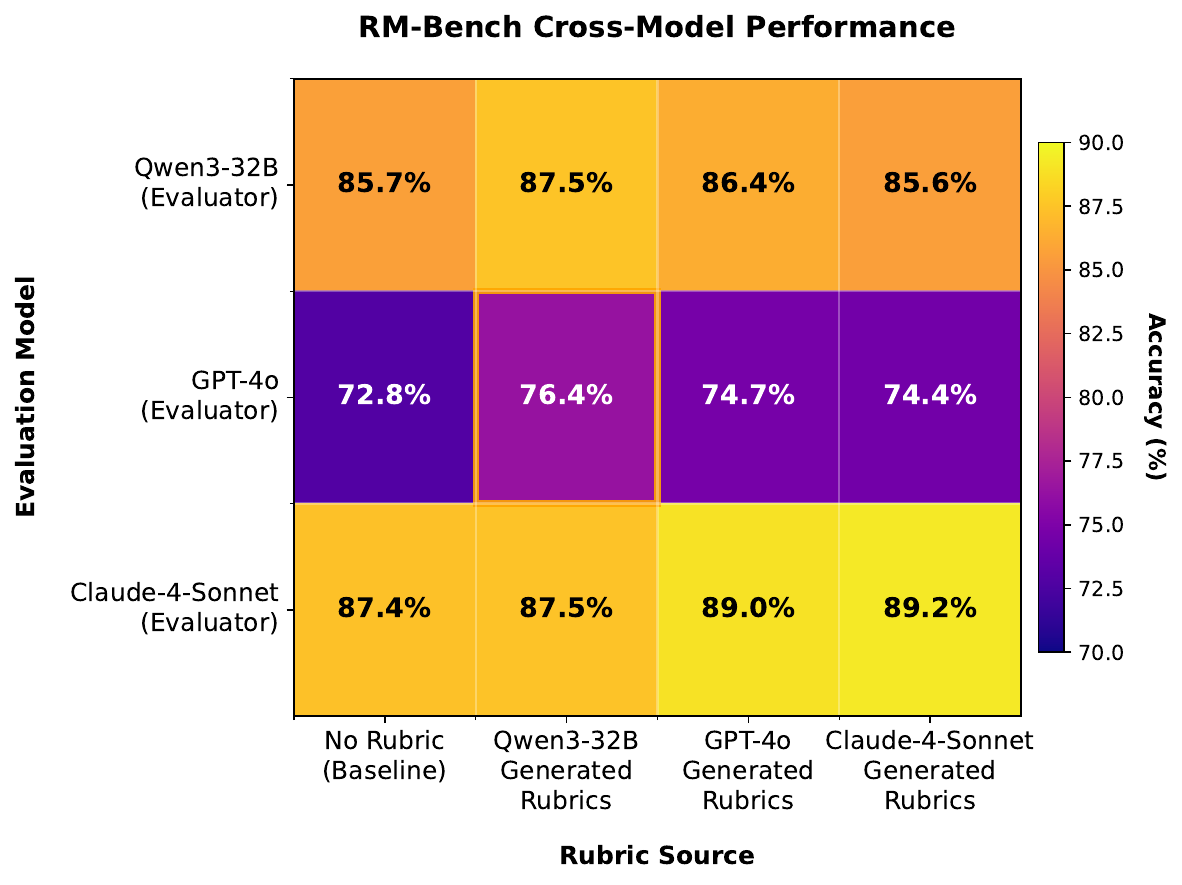}
\subcaption{RM-Bench Performance}
\label{fig:appendix_rmb_heatmap}
\end{minipage}
\caption{Cross-model rubric compatibility analysis. The heatmaps show performance when evaluation models use rubrics generated by different LLMs. The orange borders highlight the best cross-model compatibility, supporting the portability of Qwen3-32B generated rubrics.}
\label{fig:appendix_rubric_results}
\end{figure}

To select the optimal backbone for rubric generation, we analyzed cross-model compatibility of rubrics produced by three strong models: \textbf{Qwen3-32B}, \textbf{GPT-4o}, and \textbf{Claude-4-Sonnet}. Figure~\ref{fig:appendix_rubric_results} presents a heatmap where each cell shows the accuracy when a given evaluator (row) uses rubrics from a given source (column).

Applying any rubric consistently improves over the no-rubric baseline across all evaluators, confirming the general utility of explicit guidance. Moreover, \textbf{Qwen3-32B-generated rubrics exhibit the strongest cross-model compatibility on RewardBench2}, providing substantial gains even to weaker evaluators such as GPT-4o (+7.0\%). While Claude-4-Sonnet achieves the highest absolute scores as an evaluator, its generated rubrics are less effective when used by other models. This motivates our choice of Qwen3-32B as the default rubric generator.

\section{Local Induction Convergence and Test-Time Scaling Analysis}
\label{sec:appendix_query_accuracy}
\label{sec:appendix_scaling}

We analyze two key aspects of our framework: (1) the convergence behavior of verification-driven refinement during local induction, and (2) how performance scales with the voting budget during test-time inference.

\begin{figure}[htbp]
  \centering
  \begin{minipage}[b]{0.32\textwidth}
    \centering
    \includegraphics[width=\textwidth]{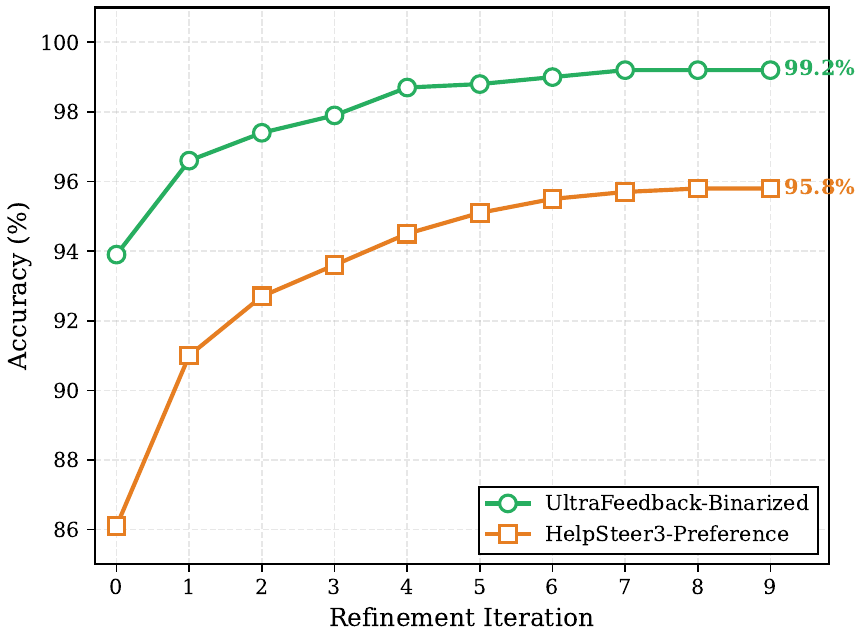}
    \subcaption{Local Induction Convergence}
    \label{fig:query_specific_accuracy}
  \end{minipage}
  \hfill
  \begin{minipage}[b]{0.32\textwidth}
    \centering
    \includegraphics[width=\textwidth]{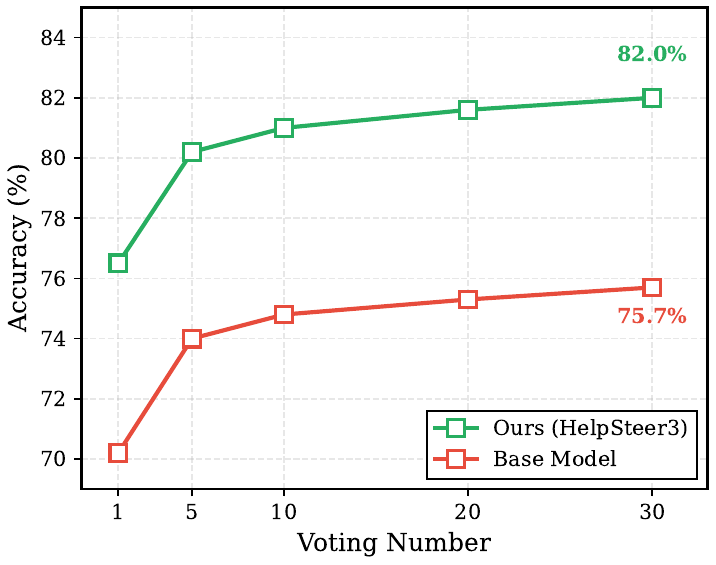}
    \subcaption{Overall Accuracy}
    \label{fig:overall_voting_accuracy}
  \end{minipage}
  \hfill
  \begin{minipage}[b]{0.32\textwidth}
    \centering
    \includegraphics[width=\textwidth]{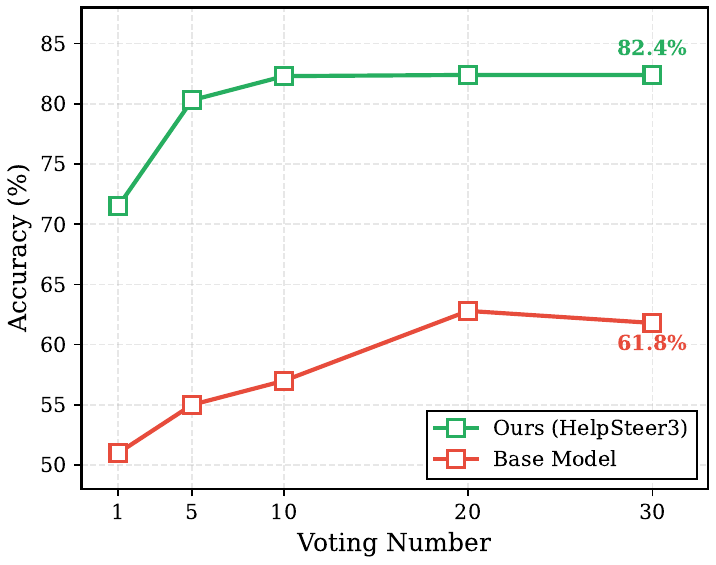}
    \subcaption{Ties Subset Accuracy}
    \label{fig:ties_voting_accuracy}
  \end{minipage}
  \caption{Convergence and scaling analysis. (a) Local induction accuracy across refinement iterations: both datasets exhibit rapid initial improvement, plateauing around iteration 6 and reaching final accuracy at iteration 9 (UltraFeedback: 99.20\%, HelpSteer3: 95.80\%). (b)--(c) Test-time scaling on RewardBench2: our rubric-guided approach (green) consistently outperforms the baseline (red) by 6--7 points overall and $\sim$25 points on ambiguous tie cases.}
  \label{fig:voting_analysis}
\end{figure}

\paragraph{Local Induction Convergence.} Figure~\ref{fig:query_specific_accuracy} tracks the accuracy of locally induced rubrics across refinement iterations. Both datasets show steep accuracy gains in the first 2--3 iterations: HelpSteer3 jumps from 86.1\% to 92.7\%, while UltraFeedback improves from 93.9\% to 97.4\%. UltraFeedback achieves higher accuracy and faster convergence than HelpSteer3 (99.20\% vs. 95.80\%), likely reflecting annotation consistency: GPT-4-based scoring exhibits more uniform patterns than human annotations. Both curves plateau around iteration 6 with marginal gains thereafter ($<$1\%), validating our early stopping mechanism.

\paragraph{Test-Time Scaling.} Figures~\ref{fig:overall_voting_accuracy}--\ref{fig:ties_voting_accuracy} examine how performance scales with the voting budget on RewardBench2. The analysis reveals three key findings:

\paragraph{Consistent Performance Advantage.} Figure~\ref{fig:overall_voting_accuracy} shows that rubric guidance maintains a stable 6--7 point advantage across all voting strategies (1, 5, 10, 20, 30). This systematic improvement indicates that the benefit from explicit rubrics is orthogonal to ensemble voting, yielding additive gains.

\paragraph{Diminishing Returns Beyond Voting@5.} Both approaches show the largest gains when scaling from Voting@1 to Voting@5, with diminishing returns thereafter. For practical deployment, Voting@5 to Voting@10 offers a reasonable trade-off between accuracy and inference cost.

\paragraph{Strong Gains on Ambiguous Cases.} Figure~\ref{fig:ties_voting_accuracy} demonstrates that rubric guidance is particularly effective on the ties subset, where the baseline struggles to make decisive judgments. The $\sim$25 point improvement highlights the discriminative power of explicit criteria in resolving ambiguous preferences.

\section{Detailed Experimental Analysis}
\label{sec:appendix_detailed_analysis}

This section provides comprehensive analyses across multiple benchmarks and evaluation dimensions, examining where our rubric-guided approach provides the largest gains.

\subsection{Cross-Domain Applicability}
\label{sec:appendix_cross_domain}

Although our rubrics are induced from general conversational preference data (HelpSteer3), they remain effective on specialized domains. Table~\ref{tab:rmbench_analysis} (Overall columns) presents cross-domain results on RM-Bench.

We observe three patterns. The \textbf{Chat} domain benefits most (+6.63\%), which is expected since the source data is conversational. \textbf{Math} (+2.60\%) and \textbf{Code} (+2.05\%) both improve despite the rubrics containing no domain-specific criteria, suggesting that general evaluation dimensions apply via semantic analogy. \textbf{Safety-Refuse} improves (+1.96\%), but Safety-Response shows a marginal decline (-0.07\%), indicating that nuanced safety scenarios may require domain-specific criteria.

\subsection{Difficulty-Stratified Analysis (RM-Bench)}

We conduct a stratified analysis on RM-Bench to understand how rubrics perform across difficulty levels (Table~\ref{tab:rmbench_analysis}). Hard samples benefit substantially more from rubric guidance (+4.68\%) compared to the overall improvement (+2.44\%). This 2x amplification on difficult cases demonstrates that rubrics provide discriminative power precisely where implicit evaluation criteria are insufficient.

Domain-specific patterns further illuminate this behavior: \textbf{Chat} shows the largest improvement on hard samples (+13.95\%), with gains also observed in \textbf{Math} (+4.54\%) and \textbf{Safety-Refuse} (+3.64\%).

\begin{table}[htbp]
\centering
\caption{Performance analysis on RM-Bench using Qwen3-32B across domains and difficulty levels.}
\label{tab:rmbench_analysis}
\footnotesize
\setlength{\tabcolsep}{5pt}
\begin{tabular}{l|ccc|ccc}
\toprule
 & \multicolumn{3}{c|}{\textbf{Overall (\%)}} & \multicolumn{3}{c}{\textbf{Hard Samples (\%)}} \\
\cmidrule(r){2-4} \cmidrule(l){5-7}
\textbf{Domain} & Base & Rubric & $\Delta$ & Base & Rubric & $\Delta$ \\
\midrule
\textbf{Overall} & 85.67 & \textbf{88.11} & +2.44 & 77.74 & \textbf{82.42} & +4.68 \\
\midrule
Chat & 70.37 & \textbf{77.00} & +6.63 & 36.95 & \textbf{50.90} & +13.95 \\
Math & 89.29 & \textbf{91.89} & +2.60 & 85.44 & \textbf{89.98} & +4.54 \\
Code & 74.90 & \textbf{76.95} & +2.05 & 72.08 & \textbf{73.54} & +1.46 \\
Safety-Refuse & 94.68 & \textbf{96.64} & +1.96 & 91.55 & \textbf{95.19} & +3.64 \\
Safety-Response & 85.42 & 85.35 & -0.07 & 68.58 & \textbf{72.61} & +4.03 \\
\bottomrule
\end{tabular}
\end{table}

\subsection{Evaluation Dimension Analysis (RewardBench2)}

Table~\ref{tab:rewardbench2_analysis} examines performance across diverse evaluation dimensions on RewardBench2. Our rubrics achieve an overall improvement of +6.72\% (from 75.55\% to 82.27\%).

The most notable finding is the improvement on the \textbf{Ties} subset (+25.49\%), jumping from 56.86\% to 82.35\%. This gain highlights the role of explicit rubrics in resolving ambiguous cases where base models struggle to make decisive judgments. The \textbf{Safety} domain also shows strong improvement (+10.34\%), suggesting rubric guidance helps navigate nuanced safety considerations.

\begin{table}[htbp]
\centering
\caption{Performance analysis on RewardBench2 using Qwen3-32B across evaluation dimensions.}
\label{tab:rewardbench2_analysis}
\footnotesize
\setlength{\tabcolsep}{8pt}
\begin{tabular}{lccc}
\toprule
\textbf{Dimension} & \textbf{Base (\%)} & \textbf{Rubric (\%)} & \textbf{$\Delta$ (\%)} \\
\midrule
\textbf{Overall} & 75.55 & \textbf{82.27} & +6.72 \\
\midrule
Precise IF & 46.88 & \textbf{52.50} & +5.62 \\
Ties & 56.86 & \textbf{82.35} & +25.49 \\
Factuality & 65.68 & \textbf{74.53} & +8.84 \\
Focus & 92.93 & \textbf{93.13} & +0.20 \\
Safety & 77.13 & \textbf{87.47} & +10.34 \\
Math & 85.79 & \textbf{86.34} & +0.55 \\
\bottomrule
\end{tabular}
\end{table}

\section{Analysis of Core Rubrics}
\label{sec:appendix_rubric_diagnostics}

We analyze the induced rubric set from HelpSteer3 through both quantitative metrics and qualitative case studies.

\subsection{Quantitative Analysis}

To validate that our framework produces high-quality evaluation criteria, we apply the analysis framework defined in Section~\ref{sec:rubric_analysis}. Table~\ref{tab:rubric_diagnostics} quantifies the utility of each rubric dimension.

Foundational dimensions like ``Clarity and Structured Presentation'' exhibit very high coverage (97.92\%) and substantial contribution (7.09\% accuracy drop if removed), serving as the backbone of evaluation. In contrast, specialized dimensions like ``Narrative and Context Fidelity'' have lower coverage (71.91\%) but the highest precision (68.24\%), providing discriminative power in scenarios that broader dimensions might miss. The positive contribution scores across all dimensions confirm that our information-theoretic selection yields a complementary set where each element adds measurable value. Complete rubric collections are presented in Appendix~\ref{sec:rubric_collections}.

\begin{table}[h]
\caption{Analysis of the final rubric dimensions induced from HelpSteer3.}
\label{tab:rubric_diagnostics}
\centering
\small
\setlength{\tabcolsep}{4pt}
\renewcommand{\arraystretch}{1.15}
\begin{tabular}{@{}lccc@{}}
\toprule
\textbf{Rubric Dimension} & \textbf{Coverage (\%)} & \textbf{Precision (\%)} & \textbf{Contribution ($\Delta$\%)} \\
\midrule
Factual Accuracy \& Context Consistency & 91.91 & 62.78 & 4.42 \\
Strict Compliance with Requirements & 85.90 & 59.16 & 3.72 \\
Clarity \& Structured Presentation & \textbf{97.92} & 65.07 & \textbf{7.09} \\
Comprehensive \& Coherent Coverage & 97.16 & 65.92 & 4.78 \\
Narrative \& Context Fidelity & 71.91 & \textbf{68.24} & 3.68 \\
\bottomrule
\end{tabular}
\end{table}

\subsection{Qualitative Analysis: Semantic Mapping}
\label{sec:appendix_semantic_mapping}

To explain how rubrics apply across disparate domains (e.g., from Chat to Code/Math), we examine specific cases from RM-Bench. Our analysis reveals a ``semantic mapping'' mechanism: the judge does not mechanically match keywords but instead maps abstract dimensions to domain-specific verification logic.

\paragraph{Mapping to Code.}
The general rubric \textbf{Strict Compliance with Requirements} is mapped to \emph{Function Signatures and API Contracts}. 
In a C++ implementation task (Sample ID: \texttt{code/20}), the baseline judge overlooked a subtle return type error. Guided by this rubric, the judge successfully flagged the mismatch (returning \texttt{vector<float>} instead of the requested \texttt{pair}), penalizing the response for violating the implicit interface contract.

\paragraph{Mapping to Math.}
The general rubric \textbf{Factual Accuracy and Context Consistency} is mapped to \emph{Calculation Precision and Derivation Rigor}.
In a mathematical reasoning task (Sample ID: \texttt{math/2655}), the judge penalized ``premature rounding'' in intermediate steps. While the final answer was approximately correct, the rubric guidance led the judge to critique process fidelity, ensuring strict precision requirements.

\paragraph{Mapping to Conversational Quality.}
The general rubric \textbf{Narrative and Context Fidelity} is mapped to \emph{Contextual Appropriateness and Detail Preservation}.
In a recipe request scenario (Sample ID: \texttt{chat/100}), the judge used this rubric to evaluate whether the response preserved culturally authentic details (e.g., using ``sweet soy sauce'' vs. generic ``soy sauce'' for Indonesian cuisine), demonstrating how general rubrics can capture domain-specific nuances through semantic analogy.

\section{Multi-Seed Stability Analysis}
\label{sec:appendix_multiseed}

To validate the stability of our approach against the inherent stochasticity of LLM inference, we conducted multi-seed experiments. We ran the complete rubric induction pipeline 5 independent times with different random seeds, sampling distinct subsets of preference data from HelpSteer3. Each run produced a different rubric set, which we evaluated across all four benchmarks. Table~\ref{tab:multiseed_results} reports the mean and standard deviation.

The results confirm high stability across three dimensions:

\paragraph{Baseline Stability.} The Base models exhibit very low variance (SD $\approx 0.1\%$), confirming that the evaluation harness is stable and inference-time stochasticity has minimal impact.

\paragraph{Rubric Stability.} Our method has slightly higher variance (SD $\approx 0.3\%$--$0.4\%$) due to data sampling during induction, but this is negligible compared to performance gains. The consistently small standard deviations ($<0.45$) indicate that our verification-driven refinement reliably converges to high-quality rubrics regardless of the specific seed data.

\paragraph{High Signal-to-Noise Ratio.} The performance gain systematically exceeds the variance. On RewardBench2 (Qwen3-32B), the gain (+6.53\%) is approximately 19$\times$ the standard deviation (0.35\%). On RM-Bench, the gain (+2.34\%) is over 10$\times$ the standard deviation (0.22\%). Even the minimum score across all runs of ``Ours'' consistently outperforms the maximum score of ``Base,'' demonstrating that improvements are systematic rather than artifacts of random sampling.

\begin{table}[h]
\centering
\caption{Multi-seed stability analysis (Mean $\pm$ SD across 5 runs). The low variance and high signal-to-noise ratio confirm robustness.}
\label{tab:multiseed_results}
\small
\renewcommand{\arraystretch}{1.2}
\begin{tabular}{@{}llcccc@{}}
\toprule
\textbf{Model} & \textbf{Method} & \textbf{RewardBench} & \textbf{RewardBench2} & \textbf{RM-Bench} & \textbf{JudgeBench} \\
\midrule
\multirow{2}{*}{Qwen3-32B} 
& Base & 92.98 $\pm$ 0.08 & 75.62 $\pm$ 0.15 & 85.71 $\pm$ 0.11 & 75.68 $\pm$ 0.13 \\
& Ours & 93.75 $\pm$ 0.18 & 82.15 $\pm$ 0.35 & 88.05 $\pm$ 0.22 & 80.72 $\pm$ 0.31 \\
\midrule
\multirow{2}{*}{Qwen3-8B} 
& Base & 92.89 $\pm$ 0.09 & 74.42 $\pm$ 0.18 & 86.85 $\pm$ 0.12 & 73.18 $\pm$ 0.14 \\
& Ours & 93.42 $\pm$ 0.21 & 80.81 $\pm$ 0.38 & 88.15 $\pm$ 0.25 & 75.65 $\pm$ 0.29 \\
\bottomrule
\end{tabular}
\end{table}

\section{Backbone Sensitivity: Qwen vs.\ Llama}
\label{sec:appendix_backbone_sensitivity}

We investigate how backbone capability affects rubric-guided evaluation by comparing two 8B models with different instruction-following strengths: Qwen3-8B and Llama-3.1-8B. Table~\ref{tab:appendix_backbone_sensitivity} presents the results.

\begin{table}[!htbp]
\centering
\caption{Backbone sensitivity analysis. We compare zero-shot performance with rubric-guided evaluation using rubrics induced from HelpSteer3 and UltraFeedback. Values in parentheses denote improvements over the baseline.}
\label{tab:appendix_backbone_sensitivity}
\small
\setlength{\tabcolsep}{4pt}
\renewcommand{\arraystretch}{1.05}
\begin{tabular}{llcccc}
\toprule
\textbf{Backbone} & \textbf{Method} & \textbf{RewardBench} & \textbf{RewardBench2} & \textbf{RM-Bench} & \textbf{JudgeBench} \\
\midrule
\multirow{3}{*}{Qwen3-8B}
& Base & 92.93 & 74.37 & 86.90 & 73.14 \\
& Ours (HelpSteer3) & 93.50 (+0.57) & 80.91 (+6.54) & 88.28 (+1.38) & 75.71 (+2.57) \\
& Ours (UltraFeedback) & 93.10 (+0.17) & 80.54 (+6.17) & 88.60 (+1.70) & 75.43 (+2.29) \\
\midrule
\multirow{3}{*}{Llama-3.1-8B}
& Base & 76.21 & 43.80 & 55.99 & 61.71 \\
& Ours (HelpSteer3) & 79.80 (+3.59) & 45.36 (+1.56) & 58.57 (+2.58) & 63.57 (+1.86) \\
& Ours (UltraFeedback) & 79.13 (+2.92) & 46.23 (+2.43) & 59.60 (+3.61) & 64.10 (+2.39) \\
\bottomrule
\end{tabular}
\end{table}

\paragraph{Consistent Improvement Across Backbones.} Our rubrics improve both models across all benchmarks. Llama-3.1-8B shows larger relative gains (e.g., +3.59\% on RewardBench vs. +0.57\% for Qwen3-8B), suggesting that explicit rubrics provide greater benefit to weaker instruction-followers by compensating for their limited zero-shot evaluation capability.

\paragraph{Absolute Performance Bounded by Base Capability.} Despite consistent improvements, Llama-3.1-8B's absolute scores remain substantially lower than Qwen3-8B's (e.g., 46.23\% vs. 80.91\% on RewardBench2). This confirms that our framework acts as a \emph{performance multiplier} rather than a substitute for base model capability, as noted in Section~\ref{sec:limitations}.

\section{Complete Benchmark Results}
\label{sec:appendix_full_results}

\begin{table}[!htbp]
\centering
\caption{Complete Performance of Models on Four Key Benchmarks (in Percent). This table extends Table~\ref{tab:main_results_wide} with all model sizes and additional baselines.}
\label{tab:appendix_full_results}
\renewcommand{\arraystretch}{1.1} 
\scriptsize
\begin{threeparttable}
\setlength{\tabcolsep}{2.5pt} 
\begin{tabular}{ll@{\hspace{3pt}}ccccc}
\toprule
\textbf{Method Type} & \textbf{Model / Variant} & \textbf{RewardBench} & \textbf{RewardBench2} & \textbf{RM-Bench} & \textbf{JudgeBench} & \textbf{Avg.} \\
\midrule
\multicolumn{7}{@{}l}{\textit{Zero-Shot Base Models}} \\
& Qwen3-8B & 92.93 & 74.37 & 86.90 & 73.14 & 81.84 \\
& Qwen3-14B & 92.66 & 76.30 & 87.70 & 75.14 & 82.95 \\
& Qwen3-32B & 92.96 & 75.55 & 85.67 & 75.71 & 82.47 \\
& Qwen3-235B & 93.70 & 83.78 & 87.55 & 83.14 & 87.04 \\
& GPT-4o & 88.24 & 72.00 & 72.80 & 68.29 & 75.33 \\
& Claude-4-Sonnet & 94.61 & 86.70 & 85.70 & 78.29 & 86.33 \\
\midrule
\multicolumn{7}{@{}l}{\textit{In-Context Learning (k=5)}} \\
& Qwen3-8B & 90.18 & 72.57 & 86.83 & 67.71 & 79.32 \\
& Qwen3-14B & 89.58 & 74.89 & 87.29 & 70.86 & 80.66 \\
& Qwen3-32B & 90.82 & 75.24 & 85.91 & 74.00 & 81.49 \\
& Qwen3-235B & 90.42 & 81.38 & 86.91 & 82.86 & 85.39 \\
& GPT-4o & 88.89 & 73.11 & 74.02 & 68.91 & 76.23 \\
& Claude-4-Sonnet & 94.82 & 84.89 & 83.29 & 77.61 & 85.15 \\
\midrule
\multicolumn{7}{@{}l}{\textit{Training-based Reward Models}} \\
& ArmoRM-Llama3-8B-v0.1 & 90.40 & 66.50 & 69.30 & 59.70 & 71.48 \\
& J1-Llama-8B & 85.70 & {--} & 73.40 & 42.00 & {--} \\
& J1-Llama-70B & 93.30 & {--} & 82.70 & 60.00 & {--} \\
& R3-QWEN3-8B-14K & 87.50 & {--} & 82.10 & {--} & {--} \\
& R3-QWEN3-14B-LORA-4K & 89.30 & {--} & 84.90 & {--} & {--} \\
& RM-R1-Qwen-Instruct-32B & 92.90 & {--} & 79.10 & {--} & {--} \\
& RM-R1-DeepSeek-Distill-Qwen-32B & 90.90 & {--} & 83.90 & {--} & {--} \\
& Skywork-Reward-V2-Qwen3-8B & 93.70 & 78.20 & 82.60 & 73.40 & 81.98 \\
& Skywork-Reward-V2-Llama-3.1-8B-40M & 97.80 & 86.50 & 96.00 & 83.40 & 90.93 \\
\midrule
\multicolumn{7}{@{}l}{\textit{LLM-as-a-Judge: Arena-Hard Prompt}} \\
& Qwen3-8B & 85.63 & 78.93 & 85.88 & 78.57 & 82.25 \\
& Qwen3-14B & 81.41 & 84.02 & 87.07 & 79.14 & 82.91 \\
& Qwen3-32B & 89.95 & 83.22 & 87.09 & 71.43 & 82.92 \\
& Qwen3-235B & 94.44 & 86.11 & 90.50 & 82.86 & 88.48 \\
& GPT-4o & 89.08 & 74.10 & 79.56 & 68.86 & 77.90 \\
& Claude-4-Sonnet & 95.41 & 84.99 & 88.11 & 82.57 & 87.77 \\
\midrule
\multicolumn{7}{@{}l}{\textit{LLM-as-a-Judge: MT-Bench Prompt}} \\
& Qwen3-8B & 92.56 & 73.40 & 84.84 & 77.14 & 81.99 \\
& Qwen3-14B & 93.03 & 76.89 & 85.37 & 79.43 & 83.68 \\
& Qwen3-32B & 91.99 & 74.29 & 80.80 & 75.44 & 80.63 \\
& Qwen3-235B & 94.03 & 84.07 & 87.04 & 82.86 & 87.00 \\
& GPT-4o & 87.24 & 64.56 & 70.21 & 63.14 & 71.29 \\
& Claude-4-Sonnet & 95.04 & 83.27 & 83.84 & 76.57 & 84.68 \\
\midrule
\multicolumn{7}{@{}l}{\textit{LLM-as-a-Judge: ICAI}} \\
& Qwen3-8B & 93.13 & 79.73 & 88.24 & 76.71 & 84.45 \\
& Qwen3-14B & 93.33 & 83.43 & 84.87 & 80.57 & 85.55 \\
& Qwen3-32B & 93.37 & 82.57 & 87.37 & 70.85 & 83.54 \\
& Qwen3-235B & 94.34 & 87.45 & 89.67 & 85.14 & 89.15 \\
& GPT-4o & 89.61 & 75.01 & 76.79 & 63.43 & 76.21 \\
& Claude-4-Sonnet & 94.57 & 86.22 & 89.34 & 80.86 & 87.75 \\
\midrule
\multicolumn{7}{@{}l}{\textit{\textbf{Ours (HelpSteer3)}}} \\
& Qwen3-8B & 93.50 & 80.91 & 88.28 & 75.71 & 84.60 \\
& Qwen3-14B & 93.74 & 81.66 & 83.15 & 79.71 & 84.57 \\
& Qwen3-32B & 93.80 & 82.27 & 88.11 & 80.86 & 86.26 \\
& Qwen3-235B & 95.81 & 86.46 & 89.51 & 85.43 & 89.30 \\
& GPT-4o & 92.09 & 78.56 & 76.83 & 69.71 & 79.30 \\
& Claude-4-Sonnet & 95.81 & 86.90 & 89.49 & 83.18 & 88.85 \\
\midrule
\multicolumn{7}{@{}l}{\textit{\textbf{Ours (UltraFeedback)}}} \\
& Qwen3-8B & 93.10 & 80.54 & 88.60 & 75.43 & 84.42 \\
& Qwen3-14B & 93.67 & 80.91 & 88.72 & 78.86 & 85.54 \\
& Qwen3-32B & 93.03 & 80.69 & 87.50 & 79.14 & 85.09 \\
& Qwen3-235B & 94.54 & 85.97 & 89.58 & 86.29 & 89.10 \\
& GPT-4o & 90.42 & 79.00 & 76.40 & 65.71 & 77.88 \\
& Claude-4-Sonnet & 95.04 & 87.90 & 87.50 & 81.71 & 88.04 \\
\bottomrule
\end{tabular}
\begin{tablenotes}
    \item \footnotesize Scores marked with '--' are unavailable from original publications.
    \item \footnotesize Average is computed only when all four benchmark scores are available.
\end{tablenotes}
\end{threeparttable}
\end{table}

\section{Induced Rubric Collections}
\label{sec:rubric_collections}

This section presents the complete rubric sets induced by our framework from two preference datasets. Each rubric follows a \textbf{hierarchical structure}: a high-level evaluation dimension supported by concrete verification criteria.

\subsection{HelpSteer3-Preference Dataset}
\label{subsec:helpsteer3_rubrics}

\begin{tcolorbox}[
    title=Dimension 1: Factual Accuracy \& Context Consistency,
    colback=blue!5,
    colframe=blue!60,
    breakable
]
Ensure factual accuracy and consistency with established context; avoid fabrication or hallucination.
\begin{itemize}[leftmargin=*, topsep=0pt, partopsep=0pt]
    \item Verify key factual claims (names, numbers, dates, definitions, attributions) and avoid confident but unsupported statements.
    \item Maintain internal consistency: do not contradict earlier statements; ensure entities, causal relations, and constraints remain coherent throughout.
    \item In domain-knowledge or canon-sensitive queries (e.g., history, media lore, technical specs), prefer conservative, well-grounded claims; flag uncertainty rather than guessing.
    \item When information is missing or ambiguous, explicitly state assumptions, ask for clarification, or provide conditional answers instead of fabricating details.
\end{itemize}
\end{tcolorbox}

\begin{tcolorbox}[
    title=Dimension 2: Strict Compliance with Requirements,
    colback=green!5,
    colframe=green!60,
    breakable
]
Follow the user's explicit requirements, including structure, formatting, and constraints, without deviation.
\begin{itemize}[leftmargin=*, topsep=0pt, partopsep=0pt]
    \item Follow explicit constraints exactly (format, length, number of items, required keywords, forbidden content), prioritizing them over stylistic preferences.
    \item Ensure completeness: answer every sub-question and satisfy all specified deliverables (e.g., list size, sections, fields, or output schema).
    \item Match the requested output style and structure (e.g., single token vs.\ paragraph, JSON-like schema vs.\ prose, bullet list vs.\ narrative) without extra chatter.
    \item Do not introduce unstated assumptions that change the task; if requirements conflict or are unclear, call out the ambiguity and propose a minimal-resolution path.
\end{itemize}
\end{tcolorbox}

\begin{tcolorbox}[
    title=Dimension 3: Clarity \& Structured Presentation,
    colback=orange!5,
    colframe=orange!60,
    breakable
]
Prioritize clarity, conciseness, and clear structure to improve readability and directness.
\begin{itemize}[leftmargin=*, topsep=0pt, partopsep=0pt]
    \item Use clear structure (headings, numbered steps, or bullets) that mirrors the task: overview $\rightarrow$ key points $\rightarrow$ actionable next steps (when applicable).
    \item Prefer concise, non-redundant wording; remove filler and keep each sentence carrying new information.
    \item Keep terminology consistent and define uncommon terms; avoid jargon unless the user context clearly warrants it.
    \item Maintain a professional, easy-to-scan style: avoid excessive formatting, tangents, or meta commentary that distracts from the answer.
\end{itemize}
\end{tcolorbox}

\begin{tcolorbox}[
    title={Dimension 4: Comprehensive \& Coherent Coverage},
    colback=purple!5,
    colframe=purple!60,
    breakable
]
Provide comprehensive, sufficiently detailed, and coherent responses that address all parts of the prompt.
\begin{itemize}[leftmargin=*, topsep=0pt, partopsep=0pt]
    \item Cover all major aspects implied by the prompt (scope, assumptions, edge cases, and constraints) rather than addressing only the most obvious part.
    \item Provide sufficient depth: include concrete details, examples, or step-by-step reasoning where it improves understanding and usefulness.
    \item Maintain thematic coherence: keep the narrative/analysis internally aligned, with smooth transitions and a clear throughline from premise to conclusion.
    \item When trade-offs exist, make them explicit (pros/cons, alternatives, risks) and end with a succinct takeaway or recommended action.
\end{itemize}
\end{tcolorbox}

\begin{tcolorbox}[
    title=Dimension 5: Narrative \& Context Fidelity,
    colback=red!5,
    colframe=red!60,
    breakable
]
Maintain narrative and contextual fidelity by preserving tone, character dynamics, and world consistency.
\begin{itemize}[leftmargin=*, topsep=0pt, partopsep=0pt]
    \item Preserve the intended voice, tone, and persona specified or implied by the prompt; keep character behavior and speaking style consistent.
    \item Maintain continuity with the provided context (prior turns, setting, relationships, rules of the world) and avoid introducing contradictions.
    \item Match genre and audience expectations (e.g., playful vs.\ serious, formal vs.\ casual, therapeutic vs.\ comedic) while avoiding abrupt tonal shifts.
    \item In role-play or scenario-driven responses, prioritize immersive, context-aware interaction over generic checklists or out-of-character meta explanations.
\end{itemize}
\end{tcolorbox}

\subsection{UltraFeedback-Binarized Dataset}
\label{subsec:ultrafeedback_rubrics}

\begin{tcolorbox}[
    title=Dimension 1: Factual Accuracy \& Domain Grounding,
    colback=blue!5,
    colframe=blue!60,
    breakable
]
Ensure the response is factually accurate and grounded in correct domain knowledge; avoid misconceptions, logical errors, or unwarranted speculation.
\begin{itemize}[leftmargin=*, topsep=0pt, partopsep=0pt]
    \item Check that technical, scientific, and mathematical claims are correct (definitions, constraints, units, and reasoning steps).
    \item Do not accept false premises uncritically; when the prompt contains misconceptions, correct them clearly and politely.
    \item Use precise terminology and distinguish verified facts from hypotheses, opinions, or heuristics; cite sources only when requested and feasible.
    \item When key details are ambiguous or missing, ask clarifying questions or state explicit assumptions rather than guessing.
    \item For translation/paraphrase tasks, preserve meaning and factual details (entities, numbers, dates) without adding, omitting, or distorting information.
\end{itemize}
\end{tcolorbox}

\begin{tcolorbox}[
    title=Dimension 2: Explicit Requirement Fulfillment,
    colback=green!5,
    colframe=green!60,
    breakable
]
Fulfill the user's explicit requirements in structure, content, and format; adhere strictly to all stated constraints.
\begin{itemize}[leftmargin=*, topsep=0pt, partopsep=0pt]
    \item Follow prescribed structure and output schema (sections, ordering, required phrases, or templates) as specified.
    \item Respect formatting constraints (e.g., Markdown/LaTeX/code blocks, patterns, length limits) and produce valid, well-formed outputs.
    \item Address every part of multi-part requests; ensure all requested artifacts are included (examples, explanations, code, citations, etc.).
    \item Use only appropriate functions, libraries, or commands for the stated environment; avoid invalid API calls or incompatible tooling.
    \item If the task restricts sources (e.g., ``use only the provided passage''), comply strictly and avoid injecting external knowledge.
\end{itemize}
\end{tcolorbox}

\begin{tcolorbox}[
    title=Dimension 3: Clarity \& Logical Organization,
    colback=orange!5,
    colframe=orange!60,
    breakable
]
Provide clear, coherent, and logically organized reasoning that is easy to follow and verify.
\begin{itemize}[leftmargin=*, topsep=0pt, partopsep=0pt]
    \item Use step-by-step explanations or structured reasoning when helpful; make assumptions and intermediate steps explicit.
    \item Keep the response readable: correct grammar, consistent terminology, and clean formatting that matches the task.
    \item Avoid redundancy, tangents, and irrelevant details that distract from the user's goal.
    \item Make the response self-contained: provide enough context, definitions, or examples so the reader can understand without external material.
    \item For interpretation/translation, maintain fidelity by using precise connectors and preserving key relations (cause/effect, contrast, conditions).
\end{itemize}
\end{tcolorbox}

\begin{tcolorbox}[
    title=Dimension 4: Depth \& Contextual Relevance,
    colback=purple!5,
    colframe=purple!60,
    breakable
]
Demonstrate sufficient depth by providing actionable detail and tailoring the response to the user's context and constraints.
\begin{itemize}[leftmargin=*, topsep=0pt, partopsep=0pt]
    \item Include concrete, context-appropriate examples that illustrate how to apply the concepts or steps.
    \item Provide practical guidance that could be executed (e.g., implementation steps, parameters, checklists, or pseudocode where relevant).
    \item Connect abstract ideas to real-world implications and decision points; clarify when and why an approach works.
    \item Address common edge cases, pitfalls, and trade-offs; propose alternatives when constraints change.
    \item Balance breadth and depth: cover the major dimensions of the problem while offering nuanced, non-superficial analysis.
\end{itemize}
\end{tcolorbox}

\begin{tcolorbox}[
    title=Dimension 5: Ethical Responsibility \& User Alignment,
    colback=red!5,
    colframe=red!60,
    breakable
]
Prioritize ethical responsibility and align the response with the user's intent while maintaining practical usefulness.
\begin{itemize}[leftmargin=*, topsep=0pt, partopsep=0pt]
    \item Use respectful, non-harmful language; avoid amplifying offensive framing and rephrase sensitively when needed.
    \item Provide actionable help aligned with the user's goals; avoid dismissive, overly abstract, or evasive responses when safe assistance is possible.
    \item Tailor guidance to the user's stated role and context (experience level, constraints, jurisdiction, audience) without inventing personal details.
    \item Encourage constructive next steps (questions to ask, checks to perform, or follow-ups) when interaction is useful.
    \item Be transparent about uncertainty and provide brief justifications for key recommendations or conclusions when it improves trustworthiness.
\end{itemize}
\end{tcolorbox}

\section{Prompt Templates}
\label{sec:appendix_prompts}

\begin{figure}[htbp]
\centering
\begin{tcolorbox}[
    title=Local Induction Prompt,
    colback=gray!3,
    colframe=gray!60,
]
\begin{lstlisting}
Based on the following sample content and annotations, generate
{generate_number} targeted ranking rubrics.

{sample_content}

## Task Requirements
- Evaluation Mode: Pairwise (rank two responses)
- Carefully analyze the differences between responses,
  including high-quality and low-quality aspects
- Generate ranking criteria that can correctly order response quality
- Criteria should determine the relative quality order of responses
- Note: Smaller rank values indicate better quality (rank=1 is best)

## Output Format
Please output strictly in the following JSON format:
{
  "rubrics": [
    "Detailed description of the first ranking criterion",
    ...
  ],
  "reason": "Reason and basis for generating these evaluation criteria"
}
\end{lstlisting}
\end{tcolorbox}
\caption{Prompt for local induction of query-specific rubrics from a single preference pair.}
\label{fig:prompt_generate}
\end{figure}

\begin{figure}[htbp]
\centering
\begin{tcolorbox}[
    title=Pairwise Evaluation Prompt,
    colback=gray!3,
    colframe=gray!60,
]
\begin{lstlisting}
Please rank all responses based on the evaluation criteria.

Evaluation Criteria:
{rubrics}

Query: {query}

All Responses:
{responses}

## Task Requirements
- Evaluate all responses based on the evaluation criteria
- Assign a rank value to each response, smaller values indicate
  better quality (rank=1 is best)
- Keep responses in original order, only output corresponding
  rank values
- Important: No two responses can have the same rank value

## Output Format
Please output strictly in the following JSON format:
{
  "rank": [Response1_rank, Response2_rank, ...],
  "reason": "Detailed explanation of quality assessment and rank
             assignment for each response"
}
\end{lstlisting}
\end{tcolorbox}
\caption{Prompt for rubric-guided pairwise evaluation.}
\label{fig:prompt_eval}
\end{figure}

\begin{figure}[htbp]
\centering
\begin{tcolorbox}[
    title=Rubric Revision Prompt,
    colback=gray!3,
    colframe=gray!60,
]
\begin{lstlisting}
The previously generated ranking criteria failed validation.
Please generate {generate_number} improved ranking criteria
based on detailed feedback.

{sample_content}

## Previous Ranking Criteria
{rubrics}

## Detailed Validation Failure Feedback
{feedback}

## Improvement Requirements
1. Analyze Failure Reasons:
   - Why didn't the current criteria produce the correct ranking?
   - Which positions in the ranking were incorrect?
   - Were different quality levels of responses confused?

2. Key Improvement Directions:
   - Carefully analyze differences between expected and actual rankings
   - Identify relative quality levels of each response
   - Ensure criteria can accurately distinguish all quality levels

3. Criteria Development Principles:
   - Each criterion should establish a clear quality gradient
   - Criteria should consistently rank all responses
   - Criteria should cover the full quality range from best to worst
   - Criteria should be discriminative

## Output Format
Please output strictly in the following JSON format:
{
  "rubrics": [
    "Detailed description of the first improved ranking criterion",
    ...
  ],
  "reason": "Reason and basis for improving these ranking criteria"
}
\end{lstlisting}
\end{tcolorbox}
\caption{Prompt for revising rubrics based on verification feedback.}
\label{fig:prompt_revise}
\end{figure}

\begin{figure}[htbp]
\centering
\begin{tcolorbox}[
    title=Hierarchical Structuring Prompt,
    colback=gray!3,
    colframe=gray!60,
]
\begin{lstlisting}
System:
You are a professional evaluation criteria aggregation expert,
skilled at integrating multiple scattered evaluation criteria
into structured hierarchical format.

User:
Please aggregate the following evaluation suggestions into
{num_categories} or fewer structured evaluation rubrics.

## Input Evaluation Suggestions
{rubrics}

## Task Requirements
- Rubrics must be fully self-contained so that non-expert readers
  need not consult any external information
- Each rubric should assess an independent dimension and be
  non-contradictory with others
- Ensure overall judgment remains aligned and consistent

## Rubric Format
Each rubric consists of two parts:
- Dimension: A concise statement capturing the core evaluation focus
- Criteria: Multiple bullet points that provide concrete verification
  checks for the dimension

## Output Format
Please output strictly in the following JSON format:
{
  "categories": [
    {
      "dimension": "First dimension statement",
      "criteria": [
        "Concrete verification check",
        "Concrete verification check"
      ]
    },
    {
      "dimension": "Second dimension statement",
      "criteria": [
        "Concrete verification check",
        "Concrete verification check"
      ]
    }
  ],
  "reason": "Reason and basis for aggregating these rubrics"
}

Please generate the aggregated evaluation criteria.
\end{lstlisting}
\end{tcolorbox}
\caption{Prompt for organizing the core rubric set into a hierarchical structure of evaluation dimensions and concrete verification criteria.}
\label{fig:prompt_structure}
\end{figure}

\end{document}